%% file: main.tex
\documentclass{article}

\usepackage{microtype}
\usepackage{graphicx}
\usepackage{subfigure}
\usepackage{booktabs} 

\usepackage{hyperref}

\usepackage[accepted]{icml2024}

\usepackage{amsmath}
\usepackage{amssymb}
\usepackage{mathtools}
\usepackage{amsthm}

\newcommand{\ieno}{\textit{i.e.}}
\newcommand{\egno}{\textit{e.g.}}

\usepackage{graphicx}
\usepackage{enumitem}
\usepackage{multirow}
\usepackage{multicol}
\usepackage{adjustbox}
\usepackage{xcolor,colortbl}

\newcommand{\tablestyle}[2]{\setlength{\tabcolsep}{#1}\renewcommand{\arraystretch}{#2}\centering\footnotesize}
\definecolor{tabhighlight}{HTML}{e5e5e5}
\definecolor{MidnightBlue}{RGB}{25,25,112}

\usepackage[capitalize,noabbrev]{cleveref}

\theoremstyle{plain}

\theoremstyle{definition}

\theoremstyle{remark}

\usepackage[textsize=tiny]{todonotes}

\icmltitlerunning{Beyond Sole Strength: Customized Ensembles for Generalized Vision-Language Models}

\begin{document}

\twocolumn[
\icmltitle{Beyond Sole Strength: \\Customized Ensembles for Generalized Vision-Language Models}



\icmlsetsymbol{equal}{*}

\begin{icmlauthorlist}
\icmlauthor{Zhihe Lu}{yyy}
\icmlauthor{Jiawang Bai}{yyy,comp}
\icmlauthor{Xin Li}{yyy,sch}
\icmlauthor{Zeyu Xiao}{yyy,sch}
\icmlauthor{Xinchao Wang}{yyy}
\end{icmlauthorlist}


\icmlaffiliation{yyy}{National University of Singapore}
\icmlaffiliation{comp}{Tsinghua University}
\icmlaffiliation{sch}{University of Science and Technology of China}

\icmlcorrespondingauthor{Xinchao Wang}{xinchao@nus.edu.sg}

\icmlkeywords{Machine Learning, ICML}

\vskip 0.3in
]



\printAffiliationsAndNotice{}  

\begin{abstract}
Fine-tuning pre-trained vision-language models (VLMs), \egno, CLIP, for the open-world generalization has gained increasing popularity due to its practical value.
However, performance advancements are limited when relying solely on intricate algorithmic designs for a single model, even one exhibiting strong performance, \egno, CLIP-ViT-B/16.
This paper, for the first time, explores the collaborative potential of leveraging much weaker VLMs to enhance the generalization of a robust single model. 
The affirmative findings motivate us to address the generalization problem from a novel perspective, \ieno, ensemble of pre-trained VLMs. 
We introduce three customized ensemble strategies, each tailored to one specific scenario.
Firstly, we introduce the zero-shot ensemble, automatically adjusting the logits of different models based on their confidence when only pre-trained VLMs are available. 
Furthermore, for scenarios with extra few-shot samples, we propose the training-free and tuning ensemble, offering flexibility based on the availability of computing resources.
The code is available at \href{https://github.com/zhiheLu/Ensemble_VLM.git}{https://github.com/zhiheLu/Ensemble\_VLM.git}.
\end{abstract}

\section{Introduction}
\label{sec:intro}

Pre-trained large-scale vision-language models (VLMs), \egno, CLIP \cite{radford2021learning} and ALIGN \cite{jia2021scaling}, have demonstrated remarkable recognition performance on open-vocabulary downstream tasks even in a zero-shot evaluation manner.
This advancement overcomes the constraint observed in prior supervised deep models \cite{he2016deep}, which are limited to classify seen classes only, \ieno, test set should share the same label space with the training set.
Encouraged by the impressive zero-shot generalization of VLMs, recent works \cite{yu2023task,li2023graphadapter,zhou2022learning,zhou2022conditional} have explored how to adapt the VLMs for better performance on downstream tasks in an efficient tuning fashion, where only few-shot samples and limited learnable parameters are used.

\begin{figure}
    \centering   
    \includegraphics[width=0.5\textwidth]{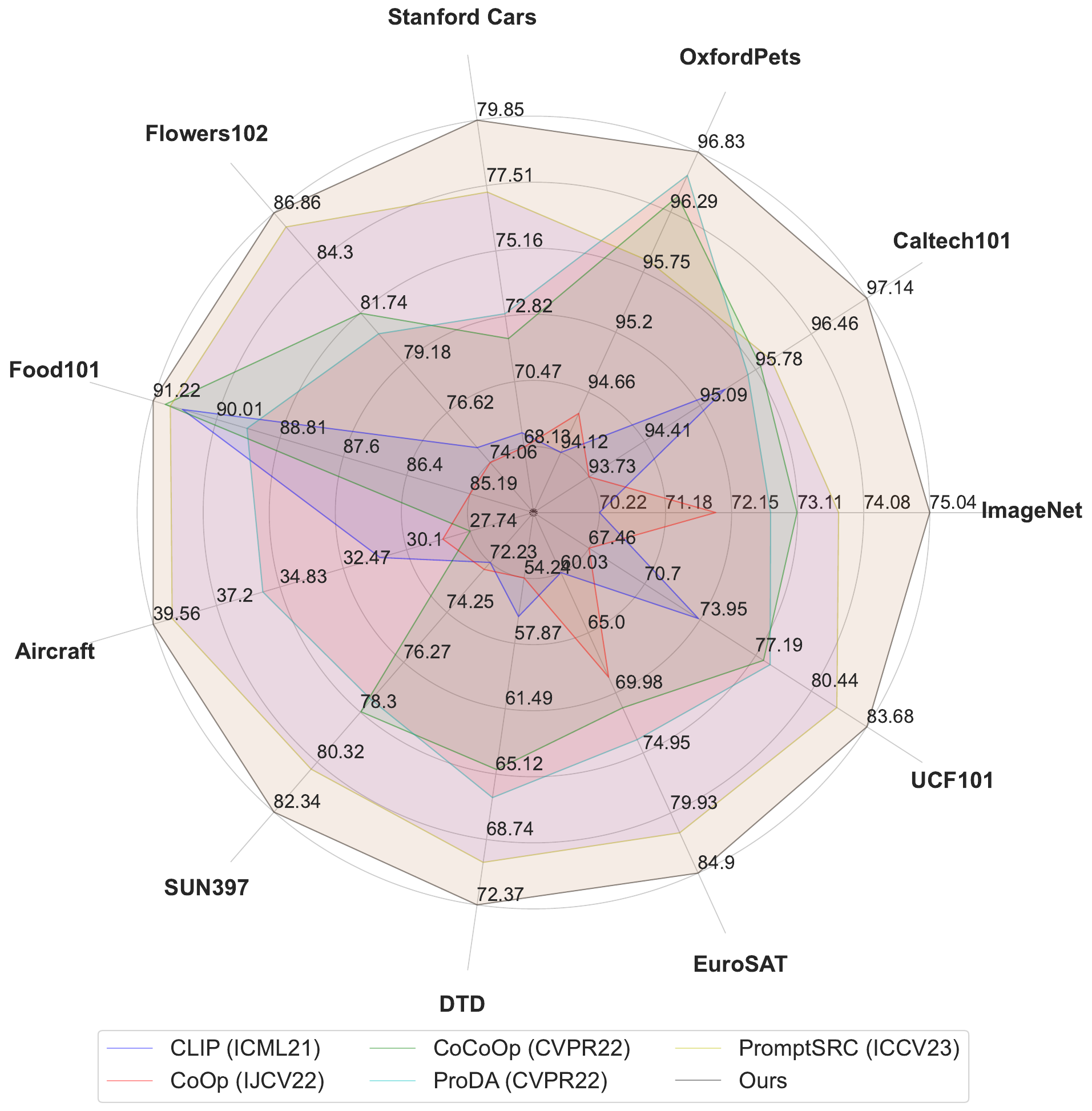}
    \caption{{\bf Comparing existing methods on base-to-new generalization}. The results indicate that the proposed method outperforms existing arts on 11 diverse datasets, often by large margins.}
    \label{fig:radar}
\end{figure}

\begin{figure*}[ht]
\centering
        \subfigure[ImageNet]{\scalebox{0.18}{\includegraphics{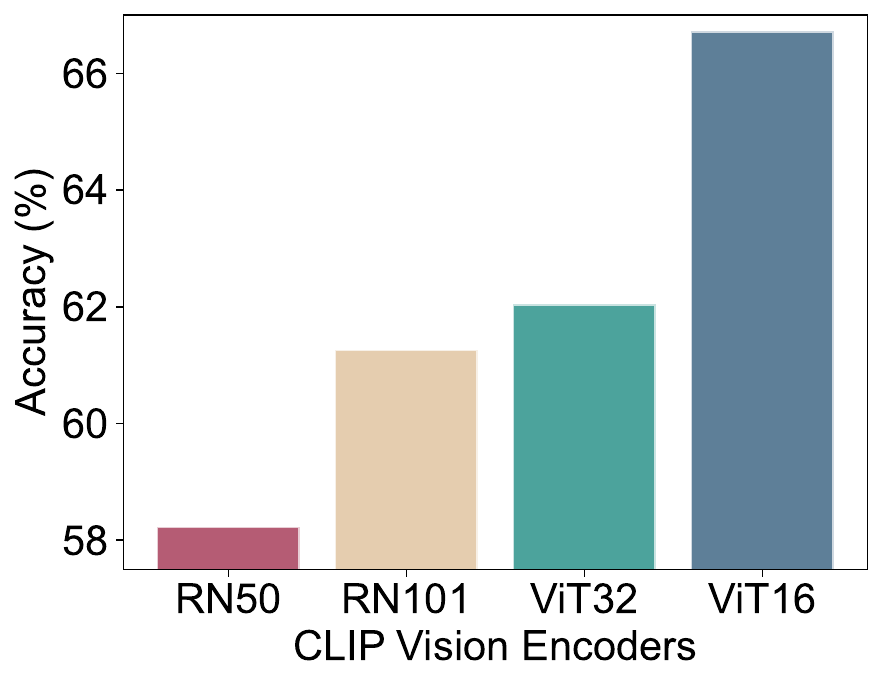}\label{fig:imagenet}}}
        \subfigure[Caltech101]{\scalebox{0.18}{\includegraphics{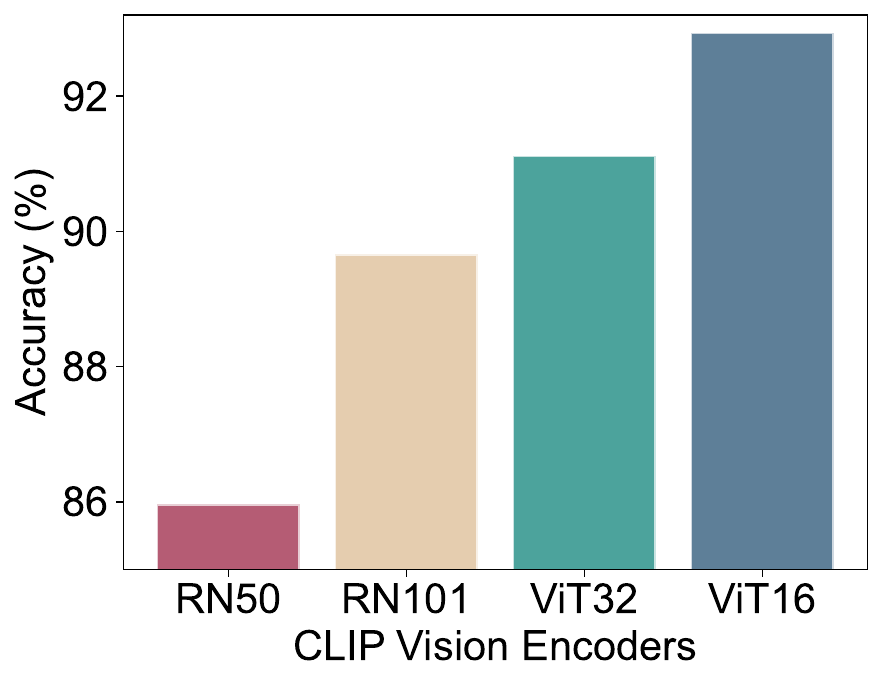}\label{fig:caltech}}}
        \subfigure[Flowers102]{\scalebox{0.18}{\includegraphics{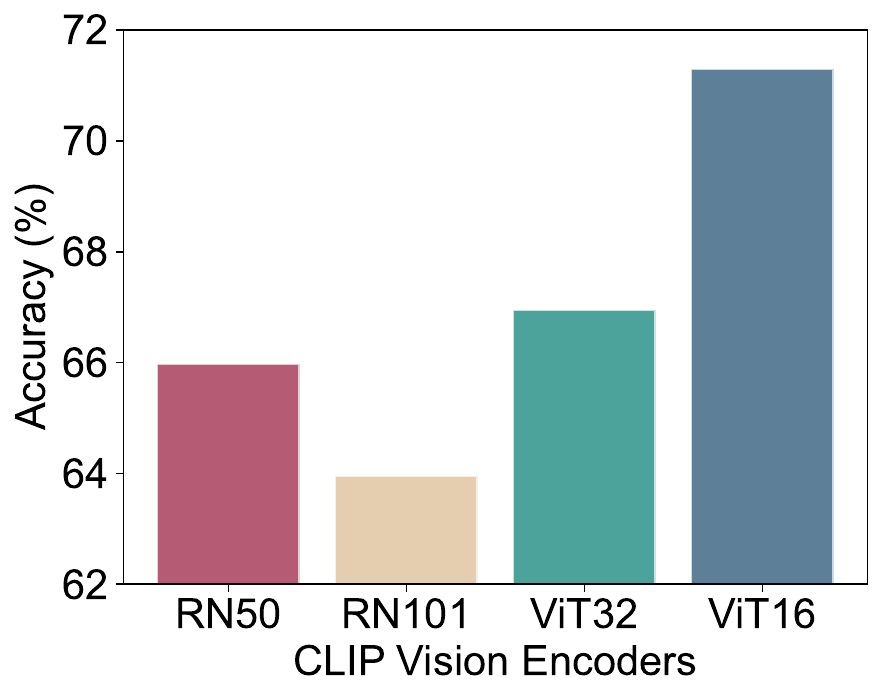}\label{fig:flower}}}
        \subfigure[DTD]{\scalebox{0.18}{\includegraphics{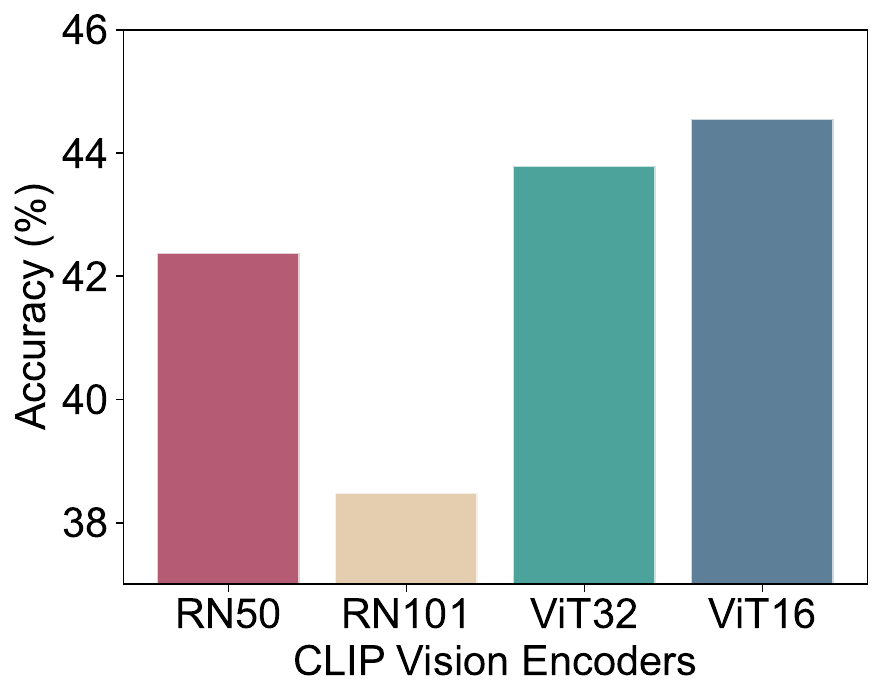}\label{fig:dtd}}}
        \subfigure[Stanford Cars]{\scalebox{0.18}{\includegraphics{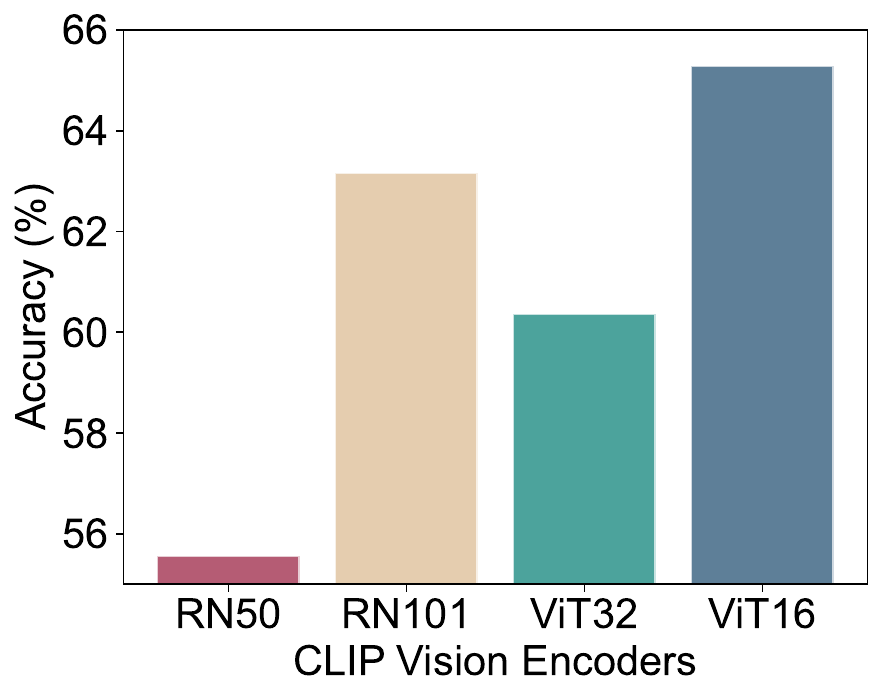}\label{fig:cars}}}
        \subfigure[Food101]{\scalebox{0.18}{\includegraphics{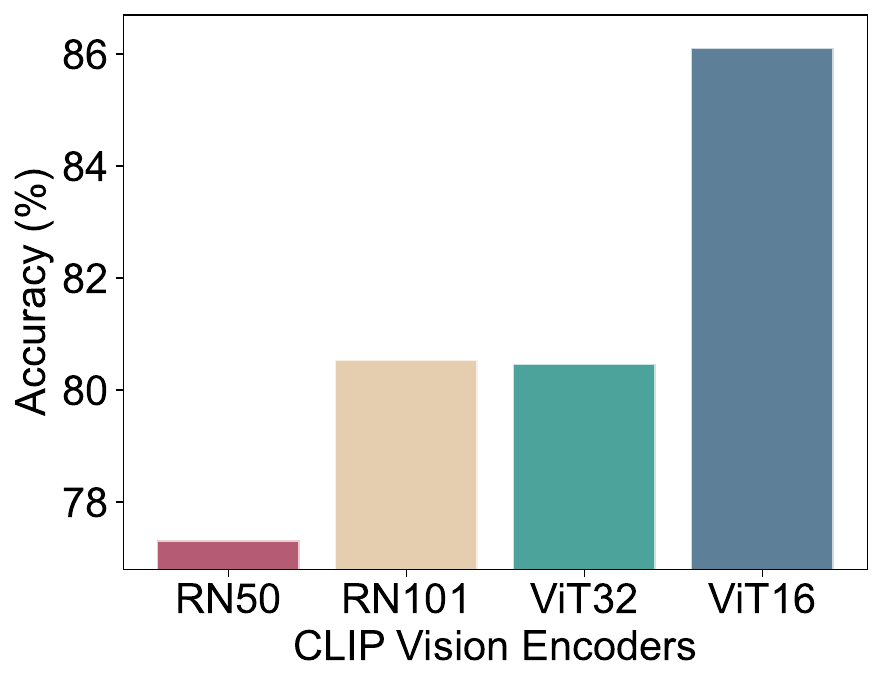}\label{fig:food}}}
  \caption{{\bf Zero-shot evaluation of pre-trained CLIP vision encoders on varying datasets}. The bar charts show that the ``weak'' models may perform better than strong ones, \egno, RN50 \textit{vs.} RN101 in (c, d) and RN101 \textit{vs.} ViT-B/32 in (e, f), encouraging us to leverage diverse models for enhanced ensemble.}
 \label{fig:zsclip}
\end{figure*}

For the efficient adaptation of VLMs, prompt learning \cite{zhou2022learning} is a popular solution. 
The concept of prompt learning was first proposed in natural language processing (NLP) \cite{shin2020autoprompt,jiang2020can,zhong2021factual}, which is to learn a prompt that dynamically suits the downstream tasks instead of prompt engineering.
CoOp \cite{zhou2022learning} was the first work to introduce prompt learning for adapting VLMs and it outperforms both handcrafted prompts and the linear probe model.
However, a critical problem of CoOp is that the learned prompt on base classes performs poorly when applied to unseen new classes even under the same data distribution.
CoCoOp \cite{zhou2022conditional} hence has been proposed to improve the performance of this base-to-new set-up by conditioning the prompt learning with the image context.
To further advance the base-to-new generalization, recent works have been proposed to leverage multi-modal prompt learning (MaPLe) \cite{khattak2023maple}, self-regulating constraints (PromptSRC) \cite{khattak2023self}, or synthesized prompts (SHIP) \cite{wang2023improving}.
It is worth noting that the default CLIP model \footnote{CLIP model specifically refers to vision encoder in this paper.} used in these methods is ViT-B/16 \cite{dosovitskiy2020vit}, one of the strongest CLIP models in Transformer structure \cite{vaswani2017attention}.
However, their performance on new classes is still not satisfactory, \egno, 70.43\% (CoCoOp, CVPR22) $\rightarrow$ 70.54\% (MaPLe, CVPR23) $\rightarrow$ 70.26\% (SHIP, ICCV23) $\rightarrow$ 70.73\% (PromptSRC, ICCV23) on ImageNet.
One possible reason is that with the advanced algorithms, the knowledge of a single model has already been fully explored for generalization, thereby leading to limited improvement.
As such, 
one question naturally arises: {\it can we leverage the prior knowledge from multiple CLIP models for better generalization even though some of them are weaker performers?}

To this end, we first investigate the zero-shot performance of four widely used CLIP models, \ieno, RN50, RN101, ViT-B/32 and ViT-B/16, on 11 diverse datasets as shown in Table \ref{tab:zero-shot}.
Intuitively, a larger model should yield better performance, but we have observed that this is not always the case across all datasets. 
For instance, the ``weak'' models may perform better than strong ones, \egno, RN50 \textit{vs.} RN101 in Figure \ref{fig:zsclip} (c, d) and RN101 \textit{vs.} ViT-B/32 in Figure \ref{fig:zsclip} (e, f).
This indicates that models with distinct architectures are inclined to predict different samples correctly, as those models may develop biases even when trained on the same dataset. 
We posit that these biases can be leveraged for improved generalization in open-world scenarios. 
Indeed, zero-shot experiments (see Table \ref{tab:ensemble_vit16}) demonstrate enhanced generalization by combining any weaker CLIP model with CLIP-ViT-B/16 while using all models performs the best.

Inspired by the favorable results of a simple ensemble, we further design three types of ensemble strategies, each tailored for a practical scenario.
{\bf First}, given solely CLIP models, we introduce the zero-shot ensemble (ZS$_{En}$), which employs confidence-aware weighting for the weaker models while preserving the dominance of the strongest model. 
This confidence-aware weighting dynamically adjusts the weights of the models' logits based on their prediction confidences of input samples.
{\bf Second}, in scenarios where additional few-shot samples ($D_{train}$) are available but without training resources, we propose the training-free ensemble (TF$_{En}$). 
This ensemble method leverages a greedy search to determine the ``optimal'' weights for model ensemble, identified when the accuracy of the ensemble is maximized on $D_{train}$.
{\bf Third}, when pre-trained models, $D_{train}$ and training resources exist simultaneously, the tuning ensemble (T$_{En}$) is proposed by learning a sample-aware weight generator (SWIG).
Specifically, the proposed SWIG takes the visual features of multiple models as input and automatically generates the ensemble weights through optimizing on the training data.

To assess the proposed methods, we employ ZS$_{En}$ for zero-shot generalization, TF$_{En}$ and T$_{En}$ for base-to-new generalization on 11 diverse datasets. 
Additionally, we examine the effectiveness of T$_{En}$ on cross-dataset evaluation.
On all tasks, our methods always yield new state-of-the-art performance, often surpassing the second best by large margins (\egno, see the results of base-to-new generalization in Figure \ref{fig:radar}).
More importantly, this work offers a new possibility for the enhanced generalization of VLMs by overcoming the limitation of existing works that rely solely on designing intricate algorithms for a single VLM.


We summarize our contributions:
\textbf{(i)} Having observed the saturated generalization performance achieved by fully exploring the knowledge from a single robust VLM, we, for the first time, propose to advance the generalization through ensemble learning.
\textbf{(ii)} We propose three ensemble strategies, each tailored for a practical situation. 
{\it First}, in cases where only CLIP models are available, we introduce the confidence-aware weighting based zero-shot ensemble (ZS$_{En}$). 
{\it Second}, when additional few-shot samples ($D_{train}$) are accessible but without extra training resources, we present the training-free ensemble based on greedy search (TF$_{En}$). 
{\it Third}, in scenarios where pre-trained models, $D_{train}$, and training resources coexist, we propose the tuning ensemble (T$_{En}$) by incorporating a sample-aware weight generator (SWIG).
\textbf{(iii)} We assess ZS$_{En}$ for zero-shot generalization, achieving an average accuracy gain of {\bf 2.61\%} across 11 diverse datasets. 
Additionally, we evaluate T$_{En}$ for base-to-new and cross-dataset generalization, yielding new state-of-the-art performance.
\textbf{(iv)} This work establishes a pathway for improved generalization in VLMs, surpassing the limitations of prior methods that depend solely on crafting intricate algorithms for a single VLM.

\section{Related Work}

\subsection{Vision-Language Models}
We mainly focus on language driven visual representation learning (LDVRL) \cite{socher2013zero,frome2013devise,elhoseiny2013write,lei2015predicting,joulin2016learning,gomez2017self,li2017learning,anderson2018bottom,sariyildiz2020learning,radford2021learning,desai2021virtex}.
LDVRL based VLMs often adopt two encoders for language and vision, respectively, and are optimized by tailored matching constraints \cite{socher2013zero, gomez2017self}.
Early works employ varied approaches for language and vision modeling. 
For language, they utilize methods like unsupervised pre-trained models \cite{socher2013zero} or skip-gram text modeling \cite{frome2013devise, mikolov2013efficient, mikolov2013distributed}. 
Meanwhile, for vision, these works explore techniques like sparse coding and vector quantization \cite{coates2011importance,socher2013zero} or utilize features such as Classeme \cite{torresani2010efficient,elhoseiny2013write}.
In recent works \cite{radford2021learning, jia2021scaling, li2021supervision}, a common trend is the utilization of two deep neural networks, \egno, Transformers \cite{vaswani2017attention, dosovitskiy2020image, touvron2021training}, to independently embed language and vision inputs.
These works generally pre-train on million/billion-level image-text pairs from internet, \egno, $\sim$400M for CLIP \cite{radford2021learning} and $\sim$1B for ALIGN \cite{jia2021scaling}, employing a contrastive loss.
The resulting pre-trained VLMs showcase impressive zero-shot generalization across a range of downstream tasks.
In this paper, we endeavor to improve their zero-shot generalization by effectively leveraging the knowledge embedded in multiple models.

\subsection{Zero-Shot Generalization of VLMs}
Zero-shot generalization/learning (ZSG) \cite{chao2016empirical,xian2017zero,wang2019survey,yi2022exploring} aims to recognize novel classes by training only on base classes.
Past ZSG works often learn the compositions of attributes \cite{huynh2020fine} and word embeddings \cite{frome2013devise, jiang2020can} on base classes to tackle ``seen-class bias'' issue.
The advent of large-scale VLMs introduces novel possibilities for addressing ZSG. 
These VLMs, trained to align image-and-text pairs, can be employed for ZSG on novel classes simply by adopting prompts of those novel classes for the language encoder.

To further advance the performance of ZSG on downstream tasks, prompt learning \cite{zhou2022learning, zhou2022conditional} has been proposed to learn a prompt on base classes that exhibits better generalization on novel classes within the same data distribution.
MaPLe \cite{khattak2023maple} explores multi-model prompt learning by jointly learning hierarchical prompts at two branches of CLIP, aiming for improved generalization performance.
Recent works have introduced approaches leveraging self-regulating constraints \cite{khattak2023self} and synthesized prompts \cite{wang2023improving}. 
Despite the intricate algorithms designed, the performance improvement on ZSG remains limited. 
One potential reason is that the knowledge stored in a single model has been thoroughly explored, making further improvement challenging.
To that end, we take the initial step in exploring knowledge from multiple VLMs to enhance generalization.

\subsection{Ensemble Learning}
Ensemble learning (EL) \cite{wolpert1992stacked,breiman1996bagging,freund1996experiments,leblanc1996combining,buja2000smoothing,breiman2001random,friedman2001greedy, ha2005response,deng2012scalable,cortes2014deep,kang2020novel} is a machine learning paradigm where multiple models, often of diverse types, are combined to improve overall predictive performance and generalization.
Technically, EL methods can be broadly categorized into three main types: Bagging (Bootstrap Aggregating) \cite{breiman1996bagging,buja2000smoothing,breiman2001random,ha2005response}, Boosting \cite{freund1996experiments, friedman2001greedy, cortes2014deep} and Stacking \cite{wolpert1992stacked, leblanc1996combining, deng2012scalable, kang2020novel}.
Bagging was initially proposed in 1996 by Breiman \cite{breiman1996bagging}. The technique involves training multiple models of the same structure on diverse subsets obtained through bootstrap sampling.
Boosting \cite{freund1996experiments, friedman2001greedy, cortes2014deep} is employed in ensemble models to transform several weak learners into one with improved generalization.
In contrast, Stacking \cite{wolpert1992stacked, leblanc1996combining, deng2012scalable, kang2020novel} is an integration technique where a meta-learning model is used to combine the outputs of base models.
Recently, Cola \cite{chen2024large} leverages large language models to coordinate multiple VLMs for visual reasoning.
In this paper, without relying on other strong models, we resort to ensemble learning for better generalization of VLMs, with a specific emphasis on prediction fusion strategies in both training-free and tuning fashions.

\section{Methodology}
\label{sec:method}
In this section, we first briefly introduce the VLM employed in our method, namely CLIP \cite{radford2021learning}. 
Subsequently, we present three specific ensemble strategies designed to leverage VLMs.
The VLMs employed in this work are four widely used CLIP models: CLIP-RN50, CLIP-RN101, CLIP-ViT-B/32, and CLIP-ViT-B/16.
Note that the backbones here mainly refer to the vision encoder.

\subsection{Preliminaries}
The CLIP model is pre-trained on $\sim$400M image-text pairs with the objective of aligning two modalities within a unified embedding space using a contrastive learning loss.
This pre-training enables CLIP to effectively capture broad visual concepts and learn general visual representations.
During inference, a given image can be classified into a pre-defined category by computing the similarity between the image feature $\mathbf{f}$ extracted from the vision encoder and the textual embeddings $\{\mathbf{c}_i|i\in\{1,\dots,K\}\}$ from the language encoder.
The inputs to the language encoder are textual prompts for $K$ categories, composed of a template, \egno, ``a photo of a $\{class\}$'', and category names.
This classification approach allows CLIP to be directly applied to a new recognition task with only the category names of interest.
We formulate the inference process as follows:
\begin{equation}
    \mathbf{Logit}_i = \frac{exp(cos(\mathbf{f}, \mathbf{c}_i)/\tau)}{\sum_{j=1}^{K}exp(cos(\mathbf{f}, \mathbf{c}_i)/\tau)}
\end{equation}
, where $cos(\cdot, \cdot)$ is the cosine similarity function and $\tau$ is the learned temperature parameter.

\subsection{Zero-shot Ensemble}
\begin{figure}[ht]
    \centering
    \includegraphics[width=0.47\textwidth]{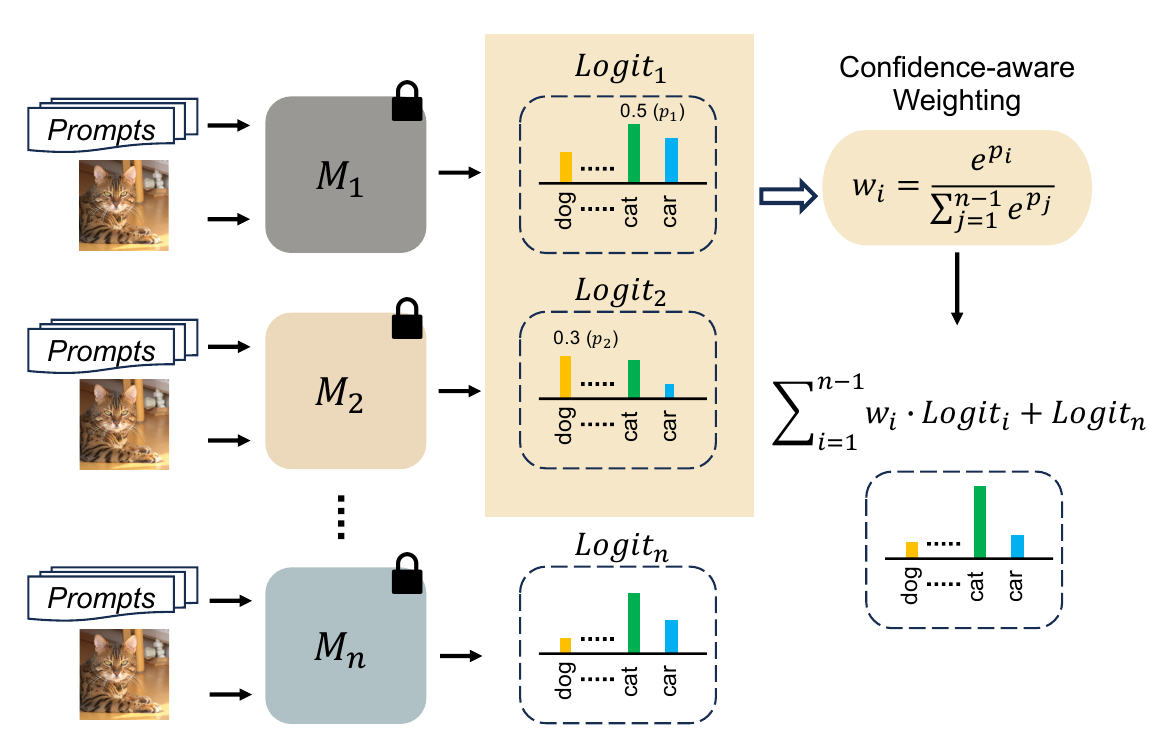}
    \caption{{\bf Illustration of our zero-shot ensemble (ZS$_{En}$)}. We assign a weight 1.0 to the best performing model, \ieno, CLIP-ViT-B/16, and use the confidence-aware weights for other VLMs.}
    \label{fig:framework_zs}
\end{figure}

The proposed zero-shot ensemble (ZS$_{En}$) aims to enhance the generalization capability of VLMs using solely pre-trained CLIP models as shown in Figure \ref{fig:framework_zs}.
Specifically, our ensemble strategy is designed with two key perspectives: (i) maintaining the dominance of the best-performing model, \ieno, CLIP-ViT-B/16; (ii) dynamically assigning weights to individual models based on their confidence for the given samples. 
The first consideration is grounded in the fact that CLIP-ViT-B/16, with inputting large patches and advanced Transformer architecture, can capture superior visual representations, leading to heightened transferability on downstream tasks. 
This is evidenced by the substantial zero-shot performance gaps observed between CLIP-ViT-B/16 and other variants. 
Consequently, preserving the dominance of CLIP-ViT-B/16 is deemed essential.
To automate the weight generation for model ensemble, we propose to use confidence-aware weights by dynamically adjusting the contribution of individual models' logits based on their prediction confidences on input samples.
The formulation of above process is defined as
\begin{align}
    \mathbf{p}_i &= max(\mathbf{Logit}_i:\{p_1, p_2, \dots, p_K\}) \\
    \mathbf{\omega}_i &= \frac{exp(\mathbf{p}_i)}{\sum_{j=1}^{n-1}exp(\mathbf{p}_j)} \\
    \mathbf{Logit}_e &= \sum_{i=1}^{n-1}\mathbf{\omega}_i \cdot \mathbf{Logit}_i + \mathbf{Logit}_n
\end{align}
, where $\mathbf{p}_i$ is the maximum prob of $\mathbf{Logit}_i$, $\mathbf{Logit}_i$ is the logit from the $i$-th VLM for an input image, $\mathbf{Logit}_n$ is the strongest VLM of n ensemble VLMs.
It is worth noting that our ZS$_{En}$ enhances the overall generalization capability of VLMs without the need for additional training data or computing resources.

\subsection{Training-free Ensemble}
\begin{figure}[ht]
    \centering
    \includegraphics[width=0.47\textwidth]{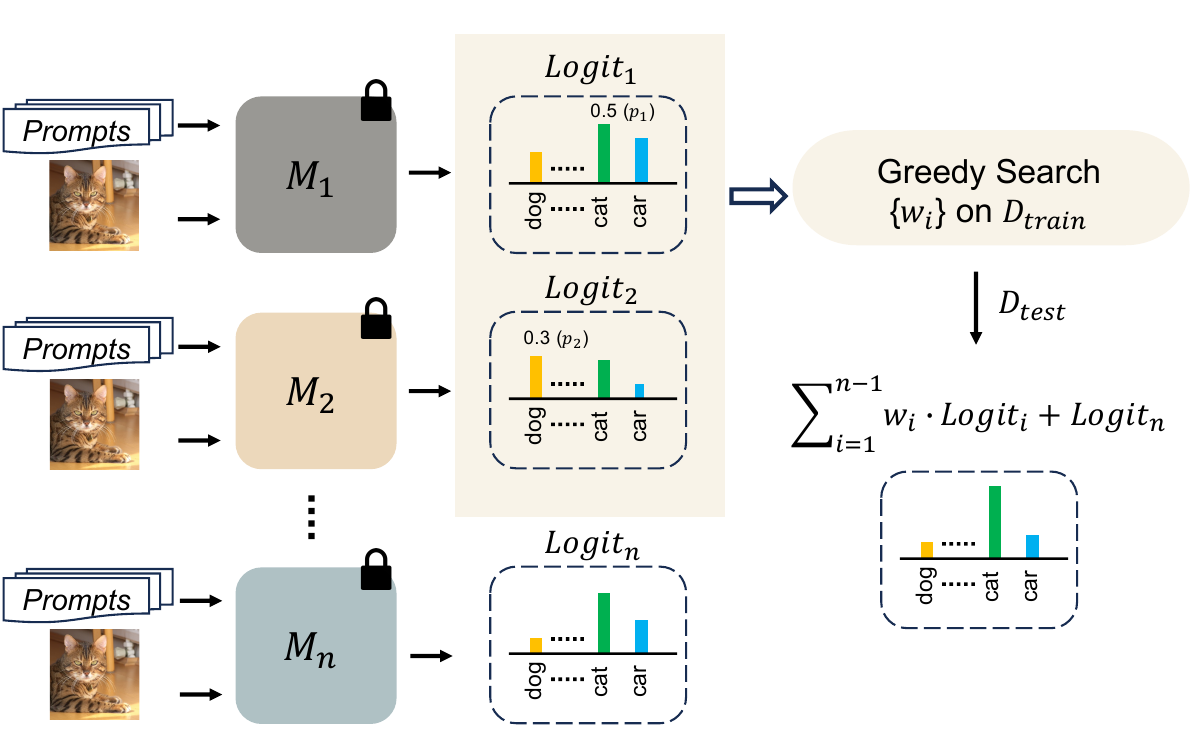}
    \caption{{\bf Illustration of our training-free ensemble (TF$_{En}$)}. We assign a weight 1.0 to the best performing model, \ieno, CLIP-ViT-B/16, and determine the weights of other VLMs by greedy searching on a given ``training'' set without training.}
    \label{fig:framework_search}
\end{figure}

We propose a training-free ensemble (see Figure \ref{fig:framework_search}) to further enhance the generalization of VLMs in scenarios where additional few-shot samples $D_{train}$ are available but there are constraints on computing resources for training.
Note that the classes in $D_{train}$ are disjoint from those in $D_{test}$ during inference, rendering this set-up both more challenging and practical.
Specifically, we implement the training-free ensemble by employing greedy search to find the optimal accuracy on a training set.
The desired weights $\{\omega_i\}$ are then determined when the optimal accuracy is achieved.
The rationale behind this approach is that samples within the same data distribution can be utilized to search for optimal parameters for classifying new classes, aligning with methodologies for base-to-new generalization \cite{zhou2022conditional, khattak2023maple, khattak2023self}.
To be specific, we formulate the above process as follows:
\begin{align}
    \{\omega_i\} &= \mathop{argmax}_{\{\omega_i\}} (\frac{1}{N} \sum_{j=1}^N \delta(y_i, \hat{y}_i) | D_{train}) \\
    \hat{y}_i &= argmax(\sum_{j=1}^{n-1} \omega_i \cdot \mathbf{Logit}_i + \mathbf{Logit}_n)
\end{align}
, where $N$ is the total number of samples in $D_{train}$, $y_i$ is the true label of the $i$-th sample, $\hat{y}_i$ is the predicted label of the $i$-th sample, $\delta(\cdot, \cdot)$ is the Kronecker delta function, which is equal to 1 if $y_i=\hat{y}_i$ and 0 otherwise.

\subsection{Tuning Ensemble}
\begin{figure}[ht]
    \centering
    \includegraphics[width=0.45\textwidth]{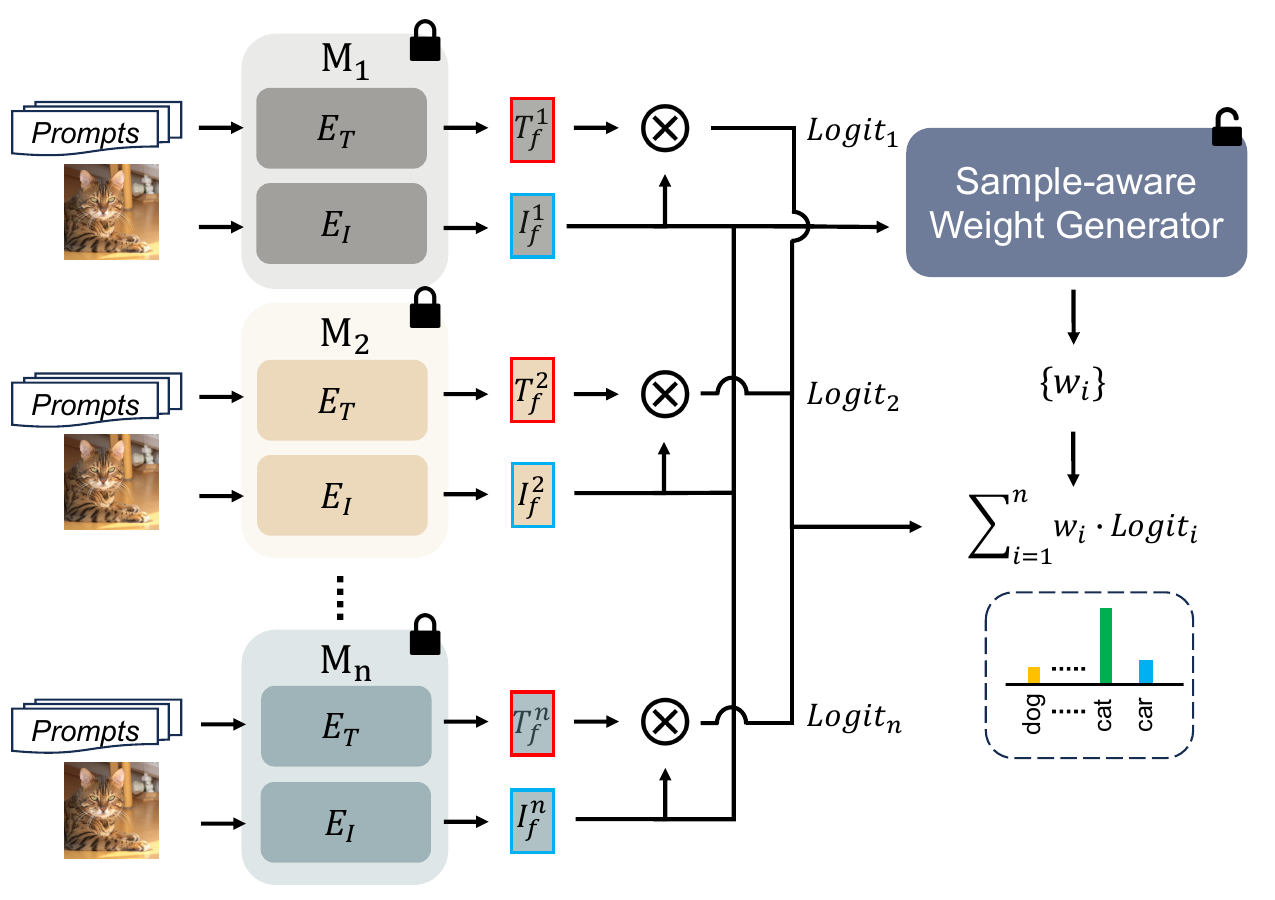}
    \caption{{\bf Illustration of our tuning ensemble (T$_{En}$)}. The proposed sample-aware weight generator (SWIG) takes sample features as input to generate sample-aware weights, which are then used for weighted prediction.}
    \label{fig:framework_weight}
\end{figure}

Following the set-up of previous works \cite{zhou2022conditional, khattak2023maple, khattak2023self}, we introduce the tuning ensemble as illustrated in Figure \ref{fig:framework_weight}.
This ensemble method employs a meta-model, sample-aware weight generator (SWIG), to learn the generation of weights on the training set $D_{train}$.
The advantage of the tuning ensemble, in comparison to the training-free version, lies in our SWIG's capability to dynamically generate weights for each sample based on the features of ensemble models. 
This facilitates test-sample-aware weight generation, with the corresponding features serving as inputs.

Concretely, to implement this tuning ensemble, we initially obtain the features $\{I_f^{1}, \dots, I_f^{n}\}$ of the input image by passing it through the vision encoders of the employed VLMs. 
Subsequently, these features are concatenated and used as the input to our proposed SWIG, which generates weights $\{\omega_1, \dots, \omega_n\}$ for the ensemble of VLMs.
These dynamically generated weights serve as a form of sample-specific attention, allowing the ensemble to adapt to the characteristics of each input image. 
This personalized weight assignment enhances the tuning ensemble's ability to tailor its predictions based on the unique features of the test samples, thereby contributing to improved generalization performance.
The aforementioned process can be formulated with the following equations:
\begin{align}
    I_f^{1}, I_f^{2}, \dots, I_f^{n} &= E_I^1(x), E_I^2(x), \dots, E_I^n(x) \\
    \{\omega_1, \dots, \omega_n\} &= \Theta_{SWIG}(concat(\{I_f^{1}, \dots, I_f^{n}\})) \\
    \mathbf{Logit}_e &= \sum_{i=1}^{n}\mathbf{\omega}_i \cdot \mathbf{Logit}_i
\end{align}
, where $E_I^i$ is the vision encoder of the $i$-th VLM, $x$ is the input image and $\Theta_{SWIG}$ is our sample-aware weight generator.

\paragraph{Remark}
\label{sec:logit}
The choice of using image features, rather than logits, as the input to our SWIG is motivated by the unique prediction mechanism of VLMs. 
VLMs calculate the cosine similarity between the image feature and textual embeddings to generate logits. 
If logits were used as the input for SWIG, the generated weights might be biased towards the textual embeddings of base classes. 
This bias is undesirable for generalization to novel classes.
By utilizing image features as the input for SWIG, we can effectively avoid this drawback. 
This approach ensures that SWIG remains sample-aware and generalizes well.

\input{Tables/zeroshot}

\input{Tables/vit16}
\input{Tables/confidence_ab}

\section{Experiments}
\label{sec:exp}
In this section, we introduce the datasets and evaluation on zero-shot, base-to-new and cross-dataset generalization.
All experiments are conducted on 11 diverse datasets, and the quantitative evaluation metric is classification accuracy. 
We also provide additional experimental results in the Appendix -- \ref{sec:appen} section.

\input{Tables/base2new}
\input{Tables/cross_dataset}
\input{Tables/swig_ab}
\subsection{Datasets}
Following prior research \cite{zhou2022learning, zhou2022conditional,khattak2023maple,khattak2023self}, we employ a set of 11 diverse datasets, covering a large range of recognition tasks.
Specifically, the benchmark comprises the following datasets: 
\textbf{(i)} ImageNet \cite{deng2009imagenet} and Caltech101 \cite{fei2004learning} for generic object classification;
\textbf{(ii)} OxfordPets \cite{parkhi2012cats}, StanfordCars \cite{krause20133d}, Flowers102 \cite{nilsback2008automated}, Food101 \cite{bossard2014food}, and FGVCAircraft \cite{maji2013fine} for fine-grained classification; 
\textbf{(iii)} SUN397 \cite{xiao2010sun} for scene recognition; 
\textbf{(iv)} UCF101 \cite{soomro2012ucf101} for action recognition; 
\textbf{(v)} DTD \cite{cimpoi2014describing} for texture classification; 
\textbf{(vi)} EuroSAT \cite{helber2019eurosat} for satellite imagery recognition.

\subsection{Implementation Details}
In our experiments, we utilize four widely used CLIP models for ensemble learning: CLIP-RN50, CLIP-RN101, CLIP-ViT-B/32, and CLIP-ViT-B/16, with CLIP-ViT-B/16 being the best performer. 
In all three ensemble strategies, the $\mathbf{Logit}$ is computed by applying the softmax function to a model's prediction.
For the training-free ensemble, the weight search space is fixed at $\{0.1, 0.2, \dots, 1.0\}$ for each VLM, excluding CLIP-ViT-B/16. 
For the tuning ensemble, we set the initial learning rate to $5e-3$ and utilize the same adjusting scheduler as in \cite{zhou2022conditional, khattak2023maple, khattak2023self}. 
The sample-aware weight generator is a two-layer MLP ($f_{dim} \rightarrow f_{dim}/32$ and $f_{dim}/32 \rightarrow num_{weight}$), which is trained for 5 epochs with a batch size of 128. 
The tuning ensemble is applied to both pre-trained CLIP models and existing baseline methods.

\subsection{Evaluation Metrics}
For zero-shot generalization, only the original test set of each dataset is used for computing accuracy since no training is needed.
In the base-to-new generalization setup, we follow \cite{zhou2022learning, zhou2022conditional, khattak2023maple, khattak2023self} to equally split the classes into two groups, \ieno, base and new classes, and then randomly sample a 16-shot training set from base classes.
The learned model is then evaluated on both base and new classes from test set.
We report three accuracy metrics: accuracy on base classes, accuracy on new classes and the harmonic mean (HM) of these two accuracies \cite{lu2023prediction}.
In contrast, for cross-dataset generalization, a model is trained on ImageNet \cite{deng2009imagenet} and then conducted cross-validation on the remaining datasets.
Note that the presented results are averaged over three runs, except for zero-shot generalization.

\subsection{Zero-shot Generalization}
We employ our zero-shot ensemble ZS$_{En}$ for zero-shot generalization as neither additional training nor extra data is needed.
The experimental results are shown in Table \ref{tab:zero-shot}.
Overall, the proposed ZS$_{En}$ can always surpass the best performing single model, \ieno, CLIP-ViT-B/16, with the averaged performance gain 2.61\% over 11 diverse datasets.
In addition, two other interesting observations can be made.

{\it First}, the strategy of ``weak helps strong'' is employed successfully in our method, where models with weaker performance contribute to the ensemble, even when there are significant performance gaps between them, such as 58.71\% (CLIP-RN50) {\it vs.} 65.27\% (CLIP-ViT-B/16). 
This is a departure from conventional ensemble methods that typically involve models with similar performance. 
The effectiveness of weaker CLIP models contributing to the ensemble and aiding stronger ones can be attributed to the diverse structures and training on a large number of image-text pairs. 
While weaker models may have lower individual performance, they capture unique aspects of the data distribution and provide complementary information. 
The ensemble leverages this diversity, allowing the collective strength of the models to cover a broader range of patterns and features present in the data, thereby enhancing overall generalization performance. 

{\it Second}, we found that the zero-shot performance of CLIP models with ResNet-based backbones does not consistently adhere to the expectation that larger models outperform smaller ones. 
On 5 out of 11 datasets, CLIP-RN101 performs either worse or comparably to CLIP-RN50. 
This trend contrasts with the behavior of Transformer-based models, which consistently improve with larger patch sizes.
The discrepancy may be attributed to two main factors: (i) the ResNet-based vision encoder might not synergize effectively with the Transformer-based language encoder, hindering the capability of larger ResNet backbones. 
(ii) the ResNet-based structure might naturally exhibit weaker generalization capabilities compared to Transformer-based architectures.

\paragraph{Ensemble of CLIP Models}
Table \ref{tab:ensemble_vit16} presents the results of an ablation study on ensemble learning by incrementally adding a weak CLIP model to the strongest one. 
The adopted ensemble strategy involves the simple averaging of logits from the utilized models. 
The results reinforce the observation that weaker CLIP models contribute positively to the overall performance, supporting the ``weak helps strong'' phenomenon discussed earlier.

\paragraph{Effect of Confidence-aware Weighting}
The effectiveness of the proposed confidence-aware weighting (CAW) is evaluated in Table \ref{tab:con_ab}.
We compare three different ways of using CAW for model ensemble: (i) applying CAW to the three strongest models, (ii) applying CAW to all four models, and (iii) applying CAW to the three weakest models combined with CLIP ViT-B/16 (ours). 
Our experiments demonstrate that all three approaches of using CAW can improve the performance of the ensemble, with our approach achieving the best results. 
This indicates that CAW is an effective technique for model ensemble, and that it is important to preserve the dominance of the best-performing model when large performance gaps exist among the ensemble models.

\subsection{Base-to-New Generalization}
We evaluate our training-free ensemble TF$_{En}$ and tuning ensemble T$_{En}$ on base-to-new generalization in Table \ref{table:base2new}.
All baseline methods use ViT-B/16 as the default vision encoder.
When employing T$_{En}$ on baseline methods, we first train them with various backbones. 
Note that CoCoOp is trained with four backbones,\ieno, RN50, RN101, ViT-B/32 and ViT-B/16 while PromptSRC is trained with two ViT-based backbones as it is specifically designed for ViT. 
Given the baseline models with diverse backbones, we then only train our T$_{En}$ for dynamic ensemble learning.

Overall, our ensemble strategies consistently demonstrate significant performance enhancements over baseline methods. 
For instance, TF$_{En}$ and T$_{En}$ improve zero-shot CLIP by 2.95\% and 3.35\%, respectively. 
CoCoOp + T$_{En}$ exhibits a performance gain of 3.60\%, while PromptSRC + T$_{En}$ achieves a gain of 1.57\%.
Another interesting finding is that our T$_{En}$ obtains the overall best performance on novel classes across 11 datasets.
This can be attributed to two reasons: (i) the pre-trained CLIP models inherently capture generalized representations, and leveraging this knowledge effectively allows for superior performance on unseen classes; (ii) existing training methods often tend to overfit on base classes without effective constraints.

\begin{figure}[htb]
\centering
        \subfigure[Base-to-new performance]{\scalebox{0.25}{\includegraphics{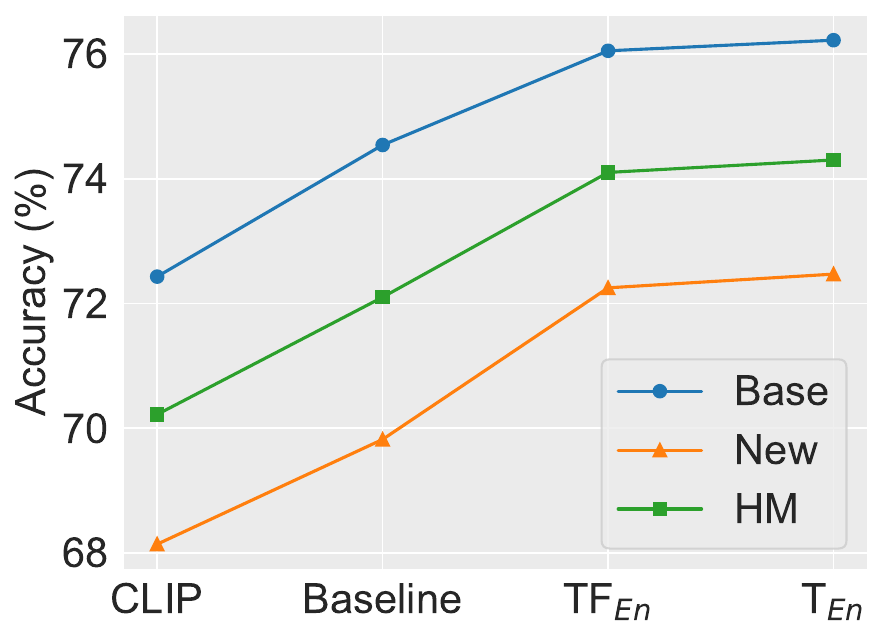}\label{fig:b2n_ab}}}
        \subfigure[Inference FPS]{\scalebox{0.25}{\includegraphics{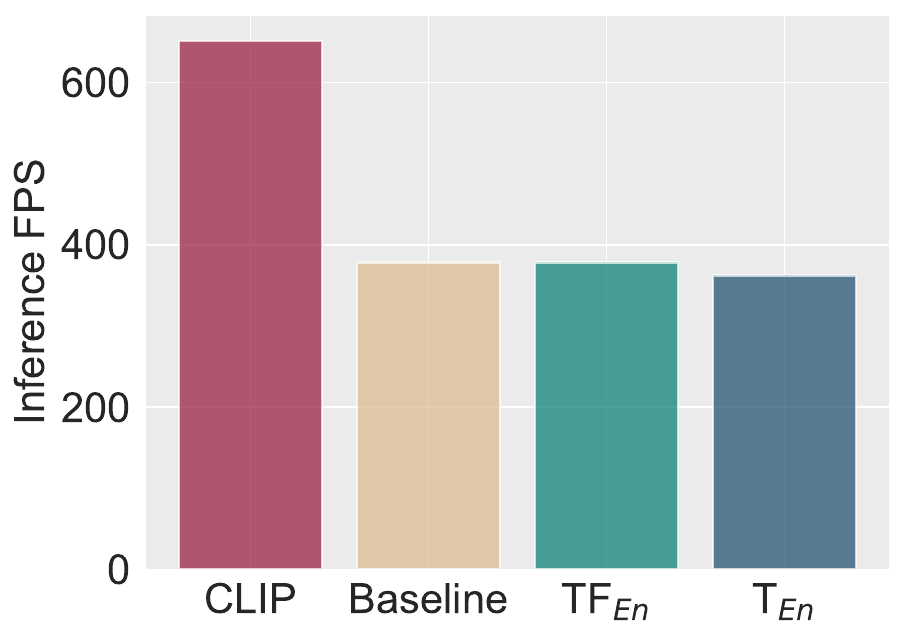}\label{fig:infer_time}}}
  \caption{{\bf Comparing our ensemble strategies to baseline in terms of performance and inference time.} Baseline uses the mean of the logits from four models as the final prediction.}
 \vspace{-2mm}
 \label{fig:base2new_ab}
\end{figure}

\paragraph{Further Analysis}

\input{Tables/dg}

\subparagraph{Ensemble Strategies}
In Figure \ref{fig:base2new_ab}, we compare the performance and inference FPS with two baselines: zero-shot CLIP-ViT-B/16 ({\bf CLIP}) and mean of the logits from four CLIP models ({\bf Baseline}). 
We found that both our TF$_{En}$ and T$_{En}$ can outperform {\bf Baseline} with the comparable inference FPS, while achieving significant improvement over {\bf CLIP} without sacrificing much inference time.

\subparagraph{Ablation Study of SWIG}
We conduct comprehensive ablation studies of our SWIG in Table \ref{tab:swig_ab}.
First, we adjust the number of SWIG's outputs to three, representing the weights for three weak VLMs, while assigning a weight of 1.0 to CLIP-ViT-B/16. 
The performance was slightly inferior to using four weights. 
This suggests that tuning our SWIG on a given dataset can generate reasonable weights for the four models, with performance comparable to maintaining the dominance of CLIP-ViT-B/16.
Second, we take the concatenated logits of four models as the input for our SWIG, but the performance is much worse as the SWIG is too biased on base classes as discussed in Sec \ref{sec:logit}. 
Finally, since SWIG consists of two linear layers: $f_{dim} \rightarrow f_{dim}/d_s$, $f_{dim}/d_s \rightarrow num_{weight}$, we ablate the downsampling scale $d_s$ and found that our SWIG is not sensitive to $d_s$.

\subsection{Cross-dataset Evaluation}
Table \ref{tab:xd} presents the results of cross-dataset evaluation for state-of-the-art methods. 
In general, our tuning ensemble demonstrates the best overall performance, surpassing other methods on 8 out of 10 datasets. 
Several noteworthy observations are discussed below. 
First, we observe that previous methods with intricate designs tend to overfit on the training dataset, leading to sub-optimal performance on other datasets. 
For instance, CoOp achieves the highest performance on ImageNet yet the lowest averaged accuracy over 10 datasets.
Second, the zero-shot CLIP model, trained on a million-level paired data, already achieves impressive performance on diverse datasets, which is slightly worse than existing SOTA methods.

\subsection{Domain Generalization}
We provide the comparison results of existing methods on domain generalization in Table \ref{tab:dg}.
Overall, the best performance is achieved by using our tuning ensemble T$_{En}$ on CoCoOp, with the accuracy gain 1.75\%.
This further indicates the importance of ensemble learning on enhanced generalization of pre-trained vision-language models.
Moreover, we found that CLIP + T$_{En}$ shows much better performance than CLIP only, but is worse than other state-of-the-art generalization methods.
This differs from the finding in the cross-dataset generalization, where CLIP + T$_{En}$ surpasses other SOTA methods by large margins.
The reason is that domain and cross-dataset generalization evaluate the model in distinct aspects.
Concretely, cross-dataset generalization is more challenging as it requires the model trained on one dataset to perform well on other datasets with unseen classes and domains, while domain generalization considers domain transfer only.
Another interesting observation is that our method can perform consistently better on both source and target domains.
This stands in stark contrast to prior methods, which often excel in the source domain but exhibit a notable decline in performance when applied to the target domain. 
For instance, methods like CoOp may outperform others in the source domain but degrade when confronted with the target domain.

\section{Conclusion}
\label{sec:con}
This paper represents the pioneering exploration of leveraging much weaker Vision-Language Models (VLMs) collaboratively to enhance the performance of a single robust one. 
Three tailored ensemble strategies are introduced, addressing distinct scenarios. Firstly, the proposed zero-shot ensemble dynamically adjusts logits based on model confidence when relying solely on pre-trained VLMs. 
Furthermore, for scenarios with additional few-shot samples, we propose both training-free and tuning ensembles, providing flexibility based on computing resource availability. 
Extensive experiments demonstrate the superiority of the proposed ensembles in diverse tasks including zero-shot evaluation and base-to-new generalization. 
Importantly, this work establishes a novel pathway for improved generalization in VLMs through ensemble learning, serving as a positive inspiration for future research endeavors.


\section*{Acknowledgments}
This project is supported by the Ministry of Education, Singapore, under its Academic Research Fund Tier 2 (Award Number: MOE-T2EP20122-0006), and the National Research Foundation, Singapore, under its Medium Sized Center for Advanced Robotics Technology Innovation.

\section*{Impact Statement}
This work pioneers the application of ensemble learning to enhance the generalization of vision-language models (VLMs) and offers three simple yet effective ensemble strategies, providing valuable insights to improve the generalization of pre-trained VLMs.
More importantly, we made several interesting observations: 
(i) the ``weak'' VLMs, displaying substantial performance gaps compared to a strong individual VLM, can serve as valuable contributors in ensemble learning.
(ii) contrary to the conventional wisdom that larger model sizes equate to better performance, we identified instances, particularly with ResNet-based \cite{he2016deep} vision encoders for CLIP \cite{radford2021learning}, where this expectation does not always hold.
(iii) Zero-shot CLIP exhibits impressive generalization capabilities on unseen scenarios. 
Optimally harnessing this power can even surpass the performance of specifically designed generalization methods.
In essence, our work provides fresh insights into the realm of ensemble learning for generalized VLMs and offers valuable findings that can contribute to advancements in the field.


\bibliography{example_paper}
\bibliographystyle{icml2024}

\newpage
\appendix
\onecolumn
\section{Appendix} \label{sec:appen}
We provide additional content to enhance the understanding of our method, which is listed as follows:
\begin{itemize}
    \item Section \ref{sec:exp_a} offers more experimental results, including the effect of our tuning ensemble on state-of-the-art methods, evaluation on other VLMs, ensembel learning with random augmentations, parameter comparison and weight pattern analysis.
    \item Section \ref{sec:limit} shows the limitations of our method, inspiring the future works.
\end{itemize}

\input{Tables/effect_en}

\input{Tables/zs_blip}

\subsection{More Experiments}
\label{sec:exp_a}

\paragraph{Effect of Tuning Ensemble}
Since we have analyzed the effect of using tuning ensemble T$_{En}$ on CLIP in main text, we further show its effect on other state-of-the-art methods in Table \ref{table:effect_en}, namely CoCoOp (CVPR22) \cite{zhou2022conditional} and PromptSRC (ICCV23) \cite{khattak2023self}.
In general, our T$_{En}$ consistently demonstrates superior performance when compared to both the baseline model and its na\"ive ensemble counterpart.
This trend underscores the efficacy of our method in effectively harnessing knowledge from multiple models.
We also found that CoCoOp + T$_{En}$ outperforms PromptSRC + T$_{En}$ in some cases, \egno, 75.49\% {\it vs.} 75.04\% on ImageNet.
This is attributed to the advantage of employing more ensemble models, showcasing the effectiveness of our T$_{En}$ in leveraging the collective power of multiple models.

\paragraph{Evaluation on Other VLMs}
We have expanded our experimental analysis to include CLIP-ViT-B/32, BLIP \cite{li2022blip} and BLIP2 \cite{li2023blip} models, applying our Zero-Shot Ensemble (ZS$_{En}$) on ImageNet, as detailed in Table \ref{tab:zs_blip}. 
It's important to clarify that this experiment does not presume prior knowledge of the best-performing model; instead, we implement confidence-aware weighting across both models indiscriminately. 
The results indicate a substantial enhancement in performance through our method, surpassing the results achieved by a simple mean ensemble.

\input{Tables/t_blip}
\input{Tables/para}
\input{Tables/weight_tf}

Our TF$_{En}$ and T$_{En}$ methods were further assessed on CLIP-ViT-B/32, BLIP and BLIP2 models, with a focus on base-to-new generalization on ImageNet. 
The results, detailed in Table \ref{tab:t_blip}, underscore the efficacy of TF$_{En}$ and T$_{En}$ across various types of VLMs.

\paragraph{Ensemble Learning with Random Augmentations}
We have carried out experiments with the highest performing single model, ViT-B/16, applying four distinct random augmentations on the ImageNet. 
The augmentations are set to include ``RandomResizedCrop(size=(224, 224), scale=(0.08, 1.0))'' and ``RandomHorizontalFlip(p=0.5)'', applied four times to each image. 
These augmented images served as inputs for the model to do emsemble learning, resulting in an accuracy of 66.81\%. 
This result slightly improves upon the baseline accuracy of 66.72\% but falls significantly short of our ZS$_{En}$ method's 70.66\% accuracy. 
The limited performance enhancement using a single model, despite varied augmentations, suggests that a singular model may not adequately capture the multifaceted aspects of an image.

\paragraph{Parameter Comparison}
We list the parameters and accuracy of baseline methods in Table \ref{tab:para}. 
From the numbers, we can see that our method achieves superior performance while preserving the nature of parameter efficiency.

\paragraph{Weight Pattern Analysis}
We have shown the searched weights of Training-Free Ensemble (TF$_{En}$) in Table \ref{tab:weight_tf}.
Note that the weight for ViT-B/16 is the constant 1.0. 
We found that the searching scheme tends to assign higher weights to the models with better zero-shot performance, evidenced on 8 out of 11 datasets. 
For models with similar performance, the weights assigned are also close.

\subsection{Limitations}
\label{sec:limit}
In this work, we take the initial stride towards incorporating ensemble learning into the generalization of VLMs. 
Our focus is on delineating potential avenues for leveraging ensemble strategies to enhance the performance of VLMs in generalized scenarios.
However, it is essential to note that our exploration, at this juncture, only scratches the surface of the capabilities that ensemble learning might offer. 
Fully harnessing the power of ensemble learning in the context of VLM generalization represents a promising avenue for future investigation.

\end{document}

%% file: Tables/zeroshot.tex
\begin{table}[htb]
\caption{{\bf Zero-shot generalization}. All compared methods are implemneted on the pre-trained CLIP \cite{radford2021learning} models without further training. Absolute gains over baseline are shown in \textcolor{MidnightBlue}{blue}.}
\centering
\label{tab:zero-shot}
\tablestyle{-12pt}{1.1}
\addtolength{\tabcolsep}{+14pt}
\resizebox{0.9\columnwidth}{!}{%
\begin{tabular}{l|c c c c | c c}
\toprule
\multirow{2}{*}{Dataset} & CLIP & CLIP & CLIP &  CLIP & ZS$_{En}$ & \multirow{2}{*}{$\Delta$}  \\
 & RN50 & RN101 & ViT-B/32 & ViT-B/16 & Ours & \\
\midrule
{\bf Average} & 58.71 & 59.52 & 61.87 & 65.27 & 67.88 &  \textcolor{MidnightBlue}{{+2.61}}\\
\midrule
ImageNet & 58.23 & 61.26 & 62.04 & 66.72 & 70.66 &  \textcolor{MidnightBlue}{{+3.94}}\\
\midrule
Caltech101 & 85.96 & 89.66 & 91.12 & 92.94 & 93.79 &  \textcolor{MidnightBlue}{{+0.85}}\\
\midrule
OxfordPets & 85.80 & 86.89 & 87.49 & 89.07 & 90.57 &  \textcolor{MidnightBlue}{{+1.50}}\\
\midrule
Stanford Cars & 55.57 & 63.16 & 60.37 & 65.29 & 70.76 &  \textcolor{MidnightBlue}{{+5.47}}\\
\midrule
Flowers102 & 65.98 & 63.95 & 66.95 & 71.30 & 73.16 &  \textcolor{MidnightBlue}{{+1.86}}\\
\midrule
Food101 & 77.32 & 80.54 & 80.47 & 86.11 & 86.78 &\textcolor{MidnightBlue}{{+0.67}}\\
\midrule
FGVC Aircraft & 17.16 & 18.18 & 19.20 & 24.87 & 25.68 & \textcolor{MidnightBlue}{{+0.81}}\\
\midrule
SUN397 & 58.53 & 58.96 & 62.00 & 62.62 & 65.91 &   \textcolor{MidnightBlue}{{+3.29}}\\
\midrule
DTD & 42.38 & 38.48 & 43.79 & 44.56 & 49.35 &  \textcolor{MidnightBlue}{{+4.79}}\\
\midrule
EuroSAT & 37.42 & 32.62 & 45.12 & 47.69 & 50.20 &  \textcolor{MidnightBlue}{{+2.51}}\\
\midrule
UCF101 & 61.43 & 61.04 & 62.07 & 66.77 & 69.84 & \textcolor{MidnightBlue}{{+3.07}}\\
\bottomrule
\end{tabular}%
}
\end{table}

%% file: Tables/vit16.tex
\begin{table}[htb]
    \caption{{\bf The ensemble of the CLIP-ViT-B/16 model with other weaker variants on ImageNet}.}
    \centering
    \label{tab:ensemble_vit16}
    \resizebox{0.9\columnwidth}{!}{
    \begin{tabular}{cccc|cc}
    \toprule
    CLIP & CLIP & CLIP & CLIP & \multirow{2}{*}{Acc} & \multirow{2}{*}{$\Delta$}  \\
    ViT-B/16 & RN50 & RN101 & ViT-B/32 & & \\
    \midrule
    $\checkmark$ & & & & 66.72 & +0.00   \\
    $\checkmark$ & $\checkmark$ & & & 67.75 &{\textcolor{MidnightBlue}{+1.03}}  \\
    $\checkmark$ & $\checkmark$ & $\checkmark$ & & 68.34 & {\textcolor{MidnightBlue}{+1.62}}  \\
    $\checkmark$ & $\checkmark$ & $\checkmark$ & $\checkmark$ & 68.76 & {\textcolor{MidnightBlue}{+2.04}}  \\
    \bottomrule
    \end{tabular}}
\end{table}


%% file: Tables/confidence_ab.tex
\begin{table}[htp]
    \caption{{\bf Evaluating confidence-aware weighting (CAW) in different manners on ImageNet}. Baseline uses the mean of the logits from four models as the final prediction.}
    \centering
    \label{tab:con_ab}
    \resizebox{0.9\columnwidth}{!}{
    \begin{tabular}{l|cc}
    \toprule
    Method & Acc & $\Delta$\\
    \midrule
    CLIP ViT-B/16 & 66.72 & +0.00 \\
    \midrule
    Baseline & 68.76 & {\textcolor{MidnightBlue}{+2.04}} \\
    CAW of 3 models (w/o RN50) & 70.01 & {\textcolor{MidnightBlue}{+3.29}} \\
    CAW of 4 models & 70.19 & {\textcolor{MidnightBlue}{+3.47}} \\
    \rowcolor{tabhighlight}
    CLIP ViT-B/16 + CAW of 3 other models & 70.66 & \textcolor{MidnightBlue}{+3.94} \\
    \bottomrule
    \end{tabular}}
\end{table}


%% file: Tables/base2new.tex
\setlength{\tabcolsep}{10.5pt}
\begin{table*}[ht]
  \caption{{\bf Comparison with state-of-the-art methods on base-to-new generalization}. All compared methods use the ViT-B/16 as the vision encoder. When employing our T$_{En}$, four of CoCoOp models, \ieno, RN50, RN101, ViT-B/32 and ViT-B/16, are utilized, while only two ViT-based models for PromptSRC as it is explicitly designed for ViT. $\dag$: reproduced by the official code. The best accuracy is in {\bf bold}.}
  \label{table:base2new}%
  \centering
  \resizebox{\textwidth}{!}{
    \begin{tabular}{ll|rrr|rrr|rrr|rrr}
    \toprule
    \multicolumn{2}{c|}{\multirow{2}[2]{*}{}} & \multicolumn{3}{c|}{\textbf{Average}} & \multicolumn{3}{c|}{\textbf{ImageNet}} & \multicolumn{3}{c|}{\textbf{Caltech101}} & \multicolumn{3}{c}{\textbf{OxfordPets}} \\
    \multicolumn{2}{c|}{} & \multicolumn{1}{c}{\textbf{Base}} & \multicolumn{1}{c}{\textbf{New}} & \multicolumn{1}{c|}{\textbf{HM}} & \multicolumn{1}{c}{\textbf{Base}} & \multicolumn{1}{c}{\textbf{New}} & \multicolumn{1}{c|}{\textbf{HM}} & \multicolumn{1}{c}{\textbf{Base}} & \multicolumn{1}{c}{\textbf{New}} & \multicolumn{1}{c|}{\textbf{HM}} & \multicolumn{1}{c}{\textbf{Base}} & \multicolumn{1}{c}{\textbf{New}} & \multicolumn{1}{c}{\textbf{HM}} \\
    \midrule
    \multicolumn{2}{l|}{CLIP~\cite{radford2021learning}} & 69.34  & 74.22  & 71.70  & 72.43  & 68.14  & 70.22  & 96.84  & 94.00  & 95.40  & 91.17  & 97.26  & 94.12  \\
    \multicolumn{2}{l|}{CoOp~\cite{zhou2022learning}} & 82.69  & 63.22  & 71.66  & 76.47  & 67.88  & 71.92  & 98.00  & 89.81  & 93.73  & 93.67  & 95.29  & 94.47  \\
    \multicolumn{2}{l|}{CoCoOp~\cite{zhou2022conditional}} & 80.47  & 71.69  & 75.83  & 75.98  & 70.43  & 73.10  & 97.96  & 93.81  & 95.84  & 95.20  & 97.69  & 96.43  \\
    \multicolumn{2}{l|}{CoCoOp$^{\dag}$~\cite{zhou2022conditional}} & 80.14 & 71.55 & 75.60 & 76.05 & 70.61 & 73.23  & 97.91 & 93.99 & 95.91 & 95.41 & 97.54 & 96.46 \\
    \multicolumn{2}{l|}{ProDA~\cite{lu2022prompt}} & 81.56  & 72.30  & 76.65  & 75.40  & 70.23  & 72.72  & 98.27  & 93.23  & 95.68  & 95.43 & 97.83  & 96.62 \\
    \multicolumn{2}{l|}{MaPLe~\cite{khattak2023maple}} & 82.28 & 75.14 & 78.55  & 76.66 & 70.54 & 73.47 & 97.74 & 94.36 & 96.02 & 95.43 & 97.76 & 96.58  \\
    \multicolumn{2}{l|}{Tip-Adapter + SHIP~\cite{wang2023improving}} & 83.80 & 76.42 & 79.94 & 77.53 & 70.26  & 73.71 & 98.32 & 94.43 & 96.34  & 94.95  & 97.09  & 96.01 \\
    \multicolumn{2}{l|}{PromptSRC~\cite{khattak2023self}} & 84.26 & 76.10 & 79.97 & 77.60 & 70.73 & 74.01 & 98.10 & 94.03 & 96.02 & 95.33 & 97.30 & 96.30 \\
    \multicolumn{2}{l|}{PromptSRC$^{\dag}$~\cite{khattak2023self}} & 83.81 & 75.69 & 79.54 & 77.57 & 70.22 & 73.71  & 98.06 & 93.96 & 95.97 & 94.59 & 97.15 & 95.85 \\
    \midrule
    \rowcolor{tabhighlight}
    \multicolumn{2}{l|}{TF$_{En}$} & 72.12 & 77.37 & 74.65 & 76.05 & 72.25 & 74.10 & 96.84 & 94.98 & 95.90 & 93.62 & 97.67 & 95.60 \\
    \rowcolor{tabhighlight}
    \multicolumn{2}{l|}{T$_{En}$} & 72.56 & {\bf 77.72} & 75.05  & 76.22 & 72.47 & 74.30 & 97.26 & 95.27 & 96.25 & 93.89 & 97.76 & 95.79 \\
    \rowcolor{tabhighlight}
    \multicolumn{2}{l|}{CoCoOp + T$_{En}$} & 83.56 & 75.27 & 79.20 & 78.39 & {\bf 72.79} & {\bf 75.49} & 98.32 & {\bf 95.74} & 97.01 & 95.69 & {\bf 98.27} & {\bf 96.96} \\
    \rowcolor{tabhighlight}
    \multicolumn{2}{l|}{PromptSRC + T$_{En}$} & {\bf 85.48} & 77.17 & {\bf 81.11} & {\bf 78.74} & 71.68 & 75.04 & {\bf 98.58} & {\bf 95.74} & {\bf 97.14} & {\bf 95.96} & 97.71 & 96.83 \\
    \bottomrule
    \end{tabular}%
  }
    \vspace{0.5mm}\\
   \resizebox{\textwidth}{!}{
    \begin{tabular}{ll|rrr|rrr|rrr|rrr}
    \toprule
    \multicolumn{2}{c|}{\multirow{2}[2]{*}{}} & \multicolumn{3}{c|}{\textbf{StanfordCars}} & \multicolumn{3}{c|}{\textbf{Flowers102}} & \multicolumn{3}{c|}{\textbf{Food101}} & \multicolumn{3}{c}{\textbf{FGVCAircraft}} \\
    \multicolumn{2}{c|}{} & \multicolumn{1}{c}{\textbf{Base}} & \multicolumn{1}{c}{\textbf{New}} & \multicolumn{1}{c|}{\textbf{HM}} & \multicolumn{1}{c}{\textbf{Base}} & \multicolumn{1}{c}{\textbf{New}} & \multicolumn{1}{c|}{\textbf{HM}} & \multicolumn{1}{c}{\textbf{Base}} & \multicolumn{1}{c}{\textbf{New}} & \multicolumn{1}{c|}{\textbf{HM}} & \multicolumn{1}{c}{\textbf{Base}} & \multicolumn{1}{c}{\textbf{New}} & \multicolumn{1}{c}{\textbf{HM}} \\
    \midrule
    \multicolumn{2}{l|}{CLIP~\cite{radford2021learning}} & 63.37  & 74.89  & 68.65  & 72.08  & 77.80  & 74.83  & 90.10  & 91.22  & 90.66  & 27.19  & 36.29 & 31.09  \\
    \multicolumn{2}{l|}{CoOp~\cite{zhou2022learning}} & 78.12  & 60.40  & 68.13  & 97.60  & 59.67  & 74.06  & 88.33  & 82.26  & 85.19  & 40.44  & 22.30  & 28.75  \\
    \multicolumn{2}{l|}{CoCoOp~\cite{zhou2022conditional}} & 70.49  & 73.59  & 72.01  & 94.87  & 71.75  & 81.71  & 90.70 & 91.29  & 90.99  & 33.41  & 23.71  & 27.74  \\
    \multicolumn{2}{l|}{CoCoOp$^{\dag}$~\cite{zhou2022conditional}} & 71.01 & 73.81 & 72.38 & 93.86 & 72.03 & 81.51 & 90.55 & 91.32 & 90.93 & 33.43 & 24.71 & 28.42 \\
    \multicolumn{2}{l|}{ProDA~\cite{lu2022prompt}} & 74.70  & 71.20  & 72.91  & 97.70  & 68.68  & 80.66  & 90.30  & 88.57  & 89.43  & 36.90  & 34.13  & 35.46 \\
    \multicolumn{2}{l|}{MaPLe~\cite{khattak2023maple}} & 72.94 & 74.00 & 73.47 & 95.92 & 72.46 & 82.56 & 90.71 & 92.05 & 91.38 & 37.44 & 35.61 & 36.50 \\
    \multicolumn{2}{l|}{Tip-Adapter + SHIP~\cite{wang2023improving}} & 79.91 & 74.62  & 77.18 & 95.35  & 77.87 & 85.73 & 90.63  & 91.51  & 91.07 & 42.62 & 35.93 & 38.99 \\
    \multicolumn{2}{l|}{PromptSRC~\cite{khattak2023self}} & 78.27 & 74.97 & 76.58 & 98.07 & 76.50 & 85.95 & 90.67 & 91.53 & 91.10 & 42.73 & 37.87 & 40.15 \\
    \multicolumn{2}{l|}{PromptSRC$^{\dag}$~\cite{khattak2023self}} & 79.19 & 75.45 & 77.27 & 97.63 & 77.07 & 86.14 & 90.35 & 91.46 & 90.90 & 42.73 & 35.63 & 38.86 \\
    \midrule
    \rowcolor{tabhighlight}
    \multicolumn{2}{l|}{TF$_{En}$} & 70.57 & 79.94 & 74.96 & 75.15 & 78.96 & 77.01 & 90.04 & 91.10 & 90.57 & 29.21 & 36.53 & 32.46 \\
    \rowcolor{tabhighlight}
    \multicolumn{2}{l|}{T$_{En}$} & 71.05 & {\bf 80.21} & 75.35  & 75.31 & {\bf 79.15} & 77.18 & 90.55 & 91.60 & 91.07 & 29.71 & {\bf 37.25} & 33.06 \\
    \rowcolor{tabhighlight}
    \multicolumn{2}{l|}{CoCoOp + T$_{En}$} & 78.69 & 78.73 & 78.71 & 97.82 & 77.23 & 86.31 & {\bf 90.93} & {\bf 92.03} & {\bf 91.48} & 36.51 & 30.17 & 33.04 \\
    \rowcolor{tabhighlight}
    \multicolumn{2}{l|}{PromptSRC + T$_{En}$} & {\bf 81.26} & 78.48 & {\bf 79.85} & {\bf 98.77} & 77.52 & {\bf 86.86} & 90.75 & 91.70 & 91.22 & {\bf 43.22} & 36.47 & {\bf 39.56} \\
    \bottomrule
    \end{tabular}%
  }
    \vspace{0.5mm}\\
    \resizebox{\textwidth}{!}{
    \begin{tabular}{ll|rrr|rrr|rrr|rrr}
    \toprule
    \multicolumn{2}{c|}{\multirow{2}[2]{*}{}} & \multicolumn{3}{c|}{\textbf{SUN397}} & \multicolumn{3}{c|}{\textbf{DTD}} & \multicolumn{3}{c|}{\textbf{EuroSAT}} & \multicolumn{3}{c}{\textbf{UCF101}} \\
    \multicolumn{2}{c|}{} & \multicolumn{1}{c}{\textbf{Base}} & \multicolumn{1}{c}{\textbf{New}} & \multicolumn{1}{c|}{\textbf{HM}} & \multicolumn{1}{c}{\textbf{Base}} & \multicolumn{1}{c}{\textbf{New}} & \multicolumn{1}{c|}{\textbf{HM}} & \multicolumn{1}{c}{\textbf{Base}} & \multicolumn{1}{c}{\textbf{New}} & \multicolumn{1}{c|}{\textbf{HM}} & \multicolumn{1}{c}{\textbf{Base}} & \multicolumn{1}{c}{\textbf{New}} & \multicolumn{1}{c}{\textbf{HM}} \\
    \midrule
    \multicolumn{2}{l|}{CLIP~\cite{radford2021learning}} & 69.36  & 75.35  & 72.23  & 53.24  & 59.90  & 56.37  & 56.48  & 64.05  & 60.03  & 70.53  & 77.50  & 73.85  \\
    \multicolumn{2}{l|}{CoOp~\cite{zhou2022learning}} & 80.60  & 65.89  & 72.51  & 79.44  & 41.18  & 54.24  & 92.19  & 54.74  & 68.69  & 84.69  & 56.05  & 67.46  \\
    \multicolumn{2}{l|}{CoCoOp~\cite{zhou2022conditional}} & 79.74  & 76.86  & 78.27  & 77.01  & 56.00  & 64.85  & 87.49  & 60.04  & 71.21  & 82.33  & 73.45  & 77.64  \\
    \multicolumn{2}{l|}{CoCoOp$^{\dag}$~\cite{zhou2022conditional}} & 79.46 & 77.24 & 78.33 & 76.01 & 54.99 & 63.81 & 86.10 & 56.49 & 68.22 & 81.73 & 74.27 & 77.82 \\
    \multicolumn{2}{l|}{ProDA~\cite{lu2022prompt}} & 78.67  & 76.93  & 77.79  & 80.67  & 56.48  & 66.44  & 83.90  & 66.00  & 73.88  & 85.23  & 71.97  & 78.04  \\
    \multicolumn{2}{l|}{MaPLe~\cite{khattak2023maple}} & 80.82 & 78.70 & 79.75 & 80.36 & 59.18 & 68.16 & 94.07 & 73.23 & 82.35 & 83.00 & 78.66 & 80.77 \\
    \multicolumn{2}{l|}{Tip-Adapter + SHIP~\cite{wang2023improving}} & 81.32 & 77.64 & 79.43 & 81.83  & 61.47 & 70.21 & 93.38  & \textbf{81.67} & \textbf{87.13} & 85.99 & 78.10 & 81.85 \\
    \multicolumn{2}{l|}{PromptSRC~\cite{khattak2023self}} & 82.67 & 78.47 & 80.52 & 83.37 & 62.97 & 71.75 & 92.90 & 73.90 & 82.32 & 87.10 & 78.80 & 82.74 \\
    \multicolumn{2}{l|}{PromptSRC$^{\dag}$~\cite{khattak2023self}} & 82.37 & 78.91 & 80.60 & 81.56 & 61.35 & 70.03 & 90.96 & 73.90 & 81.55 & 86.92 & 77.50 & 81.94 \\
    \midrule
    \rowcolor{tabhighlight}
    \multicolumn{2}{l|}{TF$_{En}$} & 72.90 & 77.08 & 74.93 & 58.72 & 64.82 & 61.62 & 56.60 & 77.85 & 65.55 & 73.63 & 79.86 & 76.62 \\
    \rowcolor{tabhighlight}
    \multicolumn{2}{l|}{T$_{En}$} & 73.18 & 77.49 & 75.27 & 59.26 & {\bf 65.10} & 62.04 & 57.69 & 78.60 & 66.54 & 74.03 & {\bf 80.03} & 76.91 \\
    \rowcolor{tabhighlight}
    \multicolumn{2}{l|}{CoCoOp + T$_{En}$} & 81.96 & 80.63 & 81.29 & 81.37 & 62.01 & 70.38 & 94.38 & 64.18 & 76.40 & 85.13 & 76.15 & 80.39 \\
    \rowcolor{tabhighlight}
    \multicolumn{2}{l|}{PromptSRC + T$_{En}$} & {\bf 83.88} & {\bf 80.86} & {\bf 82.34} & {\bf 84.72} & 63.16 & {\bf 72.37} & {\bf 95.52} & 76.41 & 84.90 & {\bf 88.83} & 79.10 & {\bf 83.68} \\
    \bottomrule
    \end{tabular}%
  }
\end{table*}%

%% file: Tables/cross_dataset.tex
\begin{table*}[ht]
    \centering
    \caption{{\bf Cross-dataset evaluation.} Our tuning ensemble T$_{En}$ achieves the overall best performance.}
    \label{tab:xd} 
    \adjustbox{max width=0.98\textwidth}{
    \begin{tabular}{l c ccccccccccc}
    \toprule
    & \textbf{Source} & \multicolumn{11}{c}{\textbf{Target}} \\ \cmidrule(lr){2-2} \cmidrule(lr){3-13}
    & \rotatebox{90}{ImageNet} & \rotatebox{90}{Caltech101} & \rotatebox{90}{OxfordPets} & \rotatebox{90}{StanfordCars} & \rotatebox{90}{Flowers102} & \rotatebox{90}{Food101} & \rotatebox{90}{Aircraft} & \rotatebox{90}{SUN397} & \rotatebox{90}{DTD} & \rotatebox{90}{EuroSAT} & \rotatebox{90}{UCF101} & \rotatebox{90}{\emph{Average}} \\
    \midrule
    CLIP \cite{radford2021learning} & 66.72 & 92.94 & 89.07 & 65.29 & 71.30 & 86.11 & 24.87 & 62.62 & 44.56 & 47.69 & 66.77 & 65.12 \\
    CoOp \cite{zhou2022learning} & {\bf 71.51} & 93.70 & 89.14 & 64.51 & 68.71 & 85.30 & 18.47 & 64.15 & 41.92 & 46.39 & 66.55 & 63.88 \\
    CoCoOp \cite{zhou2022conditional} & 71.02 & {\bf 94.43} & 90.14 & 65.32 & 71.88 & 86.06 & 22.94 & {\bf 67.36} & 45.73 & 45.37 & 68.21 & 65.74 \\
    MaPLe \cite{khattak2023maple} & 70.72 & 93.53 & 90.49 & 65.57 & 72.23 & 86.20 & 24.74 & 67.01 & 46.49 & 48.06 & 68.69 & 66.30  \\
    PromptSRC \cite{khattak2023self} & 71.27 & 93.60 & 90.25 & 65.70 & 70.25 & 86.15 & 23.90 & 67.10 & 46.87 & 45.50 & 68.75 & 65.81 \\
    \midrule
    \rowcolor{tabhighlight}
    T$_{En}$ & 70.88 & 93.91 & {\bf 90.72} & {\bf 71.94} & {\bf 72.59} & {\bf 86.68} & {\bf 26.03} & 66.07 & {\bf 49.31} & {\bf 48.18} & {\bf 69.14} & {\bf 67.46} \\
    \bottomrule
    \end{tabular}}
\end{table*}

%% file: Tables/swig_ab.tex
\begin{table*}[htb]
    \caption{{\bf Ablation Study of our SWIG} in terms of the number of output weights, input type and downsampling scales of middle layer on ImageNet.}
    \centering
    \label{tab:swig_ab}
    \resizebox{0.71\textwidth}{!}{
    \begin{tabular}{c|cc|cc|ccccc}
    \toprule
     \multirow{2}{*}{Acc} & \multicolumn{2}{c}{{\bf \# Output Weights}} & \multicolumn{2}{c}{{\bf Input Type}} & \multicolumn{5}{c}{{\bf Downsampling Scale}} \\
     \cmidrule{2-10}
     & 3 & 4 & Logits & Features & 4 & 8 & 16 & 32 & 64 \\
     \midrule
    Base & 76.11 & {\bf 76.22} & 75.45 & {\bf 76.22} & 76.16 & 76.18 & 76.16 & {\bf 76.22} & 76.18 \\
    New & 72.38 & {\bf 72.47} & 71.54 & {\bf 72.47} & 72.36 & 72.37 & 72.42 & {\bf 72.47} & 72.34 \\
    HM & 74.20 & {\bf 74.30} & 73.44 & {\bf 74.30} & 74.21 & 74.23 & 74.24 & {\bf 74.30} & 74.21 \\
    \bottomrule
    \end{tabular}}
\end{table*}

%% file: Tables/dg.tex
\begin{table*}[htb]
\caption{{\bf Comparison with existing methods in domain generalization set-up}. Our CoCoOp + T$_{En}$ shows the best performance on all target datasets.}
\label{tab:dg}
\centering
\adjustbox{max width=0.7492\textwidth}{
    \begin{tabular}{l cccccc}
    \toprule
    & \textbf{Source} & \multicolumn{5}{c}{\textbf{Target}} \\ \cmidrule(lr){2-2} \cmidrule(lr){3-7}
     & ImageNet & -V2 & -S & -A & -R & Avg. \\
    \midrule
    CLIP \cite{radford2021learning} &  66.73 & 60.83 & {46.15} & 47.77 & {73.96} & 57.18 \\
    CoOp \cite{zhou2022learning} &  71.51 & 64.20 & 47.99  & 49.71 & 75.21 & 59.28  \\
    CoCoOp \cite{zhou2022conditional} & 71.02 & {64.07} & 48.75 & 50.63 & 76.18 & 59.91 \\
    MaPLe \cite{khattak2023maple} & 70.72 & {64.07} & 49.15 & 50.90  & 76.98 & 60.28 \\
    PromptSRC \cite{khattak2023self} & 71.27 & 64.35 & 49.55 & 50.90 & 77.80 & 60.65 \\
    \midrule
    \rowcolor{tabhighlight} 
    CLIP + T$_{En}$ & 70.88 & 62.87 & 48.97 & 49.97 & 75.98 & 59.45 \\
    \rowcolor{tabhighlight}
    CoCoOp + T$_{En}$ & {\bf 73.25} & {\bf 65.73} & {\bf 50.70} & {\bf 52.11} & {\bf 78.11} & {\bf 61.66} \\
    \bottomrule
    \end{tabular}}
\end{table*}

%% file: Tables/effect_en.tex
\setlength{\tabcolsep}{10.5pt}
\begin{table*}[htb]
  \centering
  \caption{{\bf Effect of using tuning ensemble T$_{En}$ on state-of-the-art methods}. When employing our T$_{En}$, four of CoCoOp models, \ieno, RN50, RN101, ViT-B/32 and ViT-B/16, are utilized, while only two ViT-based models for PromptSRC as it is explicitly designed for ViT. $\dag$: reproduced by the official code. Mean$_{En}$: averaging the predictions of individual models. The best accuracy is in {\bf bold}.}
  \label{table:effect_en}%
  \resizebox{0.99\textwidth}{!}{
    \begin{tabular}{ll|rrr|rrr|rrr|rrr}
    \toprule
    \multicolumn{2}{c|}{\multirow{2}[2]{*}{}} & \multicolumn{3}{c|}{\textbf{Average}} & \multicolumn{3}{c|}{\textbf{ImageNet}} & \multicolumn{3}{c|}{\textbf{Caltech101}} & \multicolumn{3}{c}{\textbf{OxfordPets}} \\
    \multicolumn{2}{c|}{} & \multicolumn{1}{c}{\textbf{Base}} & \multicolumn{1}{c}{\textbf{New}} & \multicolumn{1}{c|}{\textbf{HM}} & \multicolumn{1}{c}{\textbf{Base}} & \multicolumn{1}{c}{\textbf{New}} & \multicolumn{1}{c|}{\textbf{HM}} & \multicolumn{1}{c}{\textbf{Base}} & \multicolumn{1}{c}{\textbf{New}} & \multicolumn{1}{c|}{\textbf{HM}} & \multicolumn{1}{c}{\textbf{Base}} & \multicolumn{1}{c}{\textbf{New}} & \multicolumn{1}{c}{\textbf{HM}} \\
    \midrule
    \multicolumn{2}{l|}{CoCoOp$^{\dag}$~\cite{zhou2022conditional}} & 80.14 & 71.55 & 75.60 & 76.05 & 70.61 & 73.23  & 97.91 & 93.99 & 95.91 & 95.41 & 97.54 & 96.46 \\
    \multicolumn{2}{l|}{CoCoOp + Mean$_{En}$} & 83.20 & 74.42 & 78.57 & 77.57 & 72.20 & 74.79 & 98.13 & 95.49 & 96.79 & 95.43 & 98.04 & 96.72  \\
    \rowcolor{tabhighlight}
    \multicolumn{2}{l|}{CoCoOp + T$_{En}$} & 83.56 & 75.27 & 79.20 & 78.39 & {\bf 72.79} & {\bf 75.49} & 98.32 & {\bf 95.74} & 97.01 & 95.69 & {\bf 98.27} & {\bf 96.96} \\
    \midrule
    \multicolumn{2}{l|}{PromptSRC$^{\dag}$~\cite{khattak2023self}} & 83.81 & 75.69 & 79.54 & 77.57 & 70.22 & 73.71  & 98.06 & 93.96 & 95.97 & 94.59 & 97.15 & 95.85 \\
    \multicolumn{2}{l|}{PromptSRC + Mean$_{En}$} & 85.12 & 76.63 & 80.65 & 78.14 & 71.42 & 74.63 & 98.45 & 95.25 & 96.82 & 95.55 & 97.62 & 96.57  \\
    \rowcolor{tabhighlight}
    \multicolumn{2}{l|}{PromptSRC + T$_{En}$} & {\bf 85.48} & {\bf 77.17} & {\bf 81.11} & {\bf 78.74} & 71.68 & 75.04 & {\bf 98.58} & {\bf 95.74} & {\bf 97.14} & {\bf 95.96} & 97.71 & 96.83 \\
    \bottomrule
    \end{tabular}%
  }
    \vspace{0.1cm}\\
   \resizebox{0.99\textwidth}{!}{
    \begin{tabular}{ll|rrr|rrr|rrr|rrr}
    \toprule
    \multicolumn{2}{c|}{\multirow{2}[2]{*}{}} & \multicolumn{3}{c|}{\textbf{StanfordCars}} & \multicolumn{3}{c|}{\textbf{Flowers102}} & \multicolumn{3}{c|}{\textbf{Food101}} & \multicolumn{3}{c}{\textbf{FGVCAircraft}} \\
    \multicolumn{2}{c|}{} & \multicolumn{1}{c}{\textbf{Base}} & \multicolumn{1}{c}{\textbf{New}} & \multicolumn{1}{c|}{\textbf{HM}} & \multicolumn{1}{c}{\textbf{Base}} & \multicolumn{1}{c}{\textbf{New}} & \multicolumn{1}{c|}{\textbf{HM}} & \multicolumn{1}{c}{\textbf{Base}} & \multicolumn{1}{c}{\textbf{New}} & \multicolumn{1}{c|}{\textbf{HM}} & \multicolumn{1}{c}{\textbf{Base}} & \multicolumn{1}{c}{\textbf{New}} & \multicolumn{1}{c}{\textbf{HM}} \\
    \midrule
    \multicolumn{2}{l|}{CoCoOp$^{\dag}$~\cite{zhou2022conditional}} & 71.01 & 73.81 & 72.38 & 93.86 & 72.03 & 81.51 & 90.55 & 91.32 & 90.93 & 33.43 & 24.71 & 28.42 \\
    \multicolumn{2}{l|}{CoCoOp + Mean$_{En}$} & 78.23 & 78.46 & 78.34 & 97.69 & 75.93 & 85.45 & 90.85 & 91.89 & 91.37 & 36.39 & 27.05 & 31.03 \\
    \rowcolor{tabhighlight}
    \multicolumn{2}{l|}{CoCoOp + T$_{En}$} & 78.69 & {\bf 78.73} & 78.71 & 97.82 & 77.23 & 86.31 & {\bf 90.93} & {\bf 92.03} & {\bf 91.48} & 36.51 & 30.17 & 33.04 \\
    \midrule
    \multicolumn{2}{l|}{PromptSRC$^{\dag}$~\cite{khattak2023self}} & 79.19 & 75.45 & 77.27 & 97.63 & 77.07 & 86.14 & 90.35 & 91.46 & 90.90 & 42.73 & 35.63 & 38.86 \\
    \multicolumn{2}{l|}{PromptSRC + Mean$_{En}$} & 80.80 & 78.31 & 79.54 & 98.74 & 77.07 & 86.57 & 90.56 & 91.55 & 91.05 & 42.50 & 34.17 & 37.88 \\
    \rowcolor{tabhighlight}
    \multicolumn{2}{l|}{PromptSRC + T$_{En}$} & {\bf 81.26} & 78.48 & {\bf 79.85} & {\bf 98.77} & {\bf 77.52} & {\bf 86.86} & 90.75 & 91.70 & 91.22 & {\bf 43.22} & {\bf 36.47} & {\bf 39.56} \\
    \bottomrule
    \end{tabular}%
  }
    \vspace{0.1cm}\\
    \resizebox{0.99\textwidth}{!}{
    \begin{tabular}{ll|rrr|rrr|rrr|rrr}
    \toprule
    \multicolumn{2}{c|}{\multirow{2}[2]{*}{}} & \multicolumn{3}{c|}{\textbf{SUN397}} & \multicolumn{3}{c|}{\textbf{DTD}} & \multicolumn{3}{c|}{\textbf{EuroSAT}} & \multicolumn{3}{c}{\textbf{UCF101}} \\
    \multicolumn{2}{c|}{} & \multicolumn{1}{c}{\textbf{Base}} & \multicolumn{1}{c}{\textbf{New}} & \multicolumn{1}{c|}{\textbf{HM}} & \multicolumn{1}{c}{\textbf{Base}} & \multicolumn{1}{c}{\textbf{New}} & \multicolumn{1}{c|}{\textbf{HM}} & \multicolumn{1}{c}{\textbf{Base}} & \multicolumn{1}{c}{\textbf{New}} & \multicolumn{1}{c|}{\textbf{HM}} & \multicolumn{1}{c}{\textbf{Base}} & \multicolumn{1}{c}{\textbf{New}} & \multicolumn{1}{c}{\textbf{HM}} \\
    \midrule
    \multicolumn{2}{l|}{CoCoOp$^{\dag}$~\cite{zhou2022conditional}} & 79.46 & 77.24 & 78.33 & 76.01 & 54.99 & 63.81 & 86.10 & 56.49 & 68.22 & 81.73 & 74.27 & 77.82 \\
    \multicolumn{2}{l|}{CoCoOp + Mean$_{En}$} & 81.61 & 80.41 & 81.01 & 81.02 & 61.15 & 69.70 & 93.77 & 62.15 & 74.75 & 84.54 & 75.88 & 79.98 \\
    \rowcolor{tabhighlight}
    \multicolumn{2}{l|}{CoCoOp + T$_{En}$} & 81.96 & 80.63 & 81.29 & 81.37 & 62.01 & 70.38 & 94.38 & 64.18 & 76.40 & 85.13 & 76.15 & 80.39 \\
    \midrule
    \multicolumn{2}{l|}{PromptSRC$^{\dag}$~\cite{khattak2023self}} & 82.37 & 78.91 & 80.60 & 81.56 & 61.35 & 70.03 & 90.96 & 73.90 & 81.55 & 86.92 & 77.50 & 81.94 \\
    \multicolumn{2}{l|}{PromptSRC + Mean$_{En}$} & 83.65 & 80.36 & 81.97 & 84.38 & 62.72 & 71.96 & 95.09 & 76.02 & 84.49 & 88.42 & 78.46 & 83.14 \\
    \rowcolor{tabhighlight}
    \multicolumn{2}{l|}{PromptSRC + T$_{En}$} & {\bf 83.88} & {\bf 80.86} & {\bf 82.34} & {\bf 84.72} & {\bf 63.16} & {\bf 72.37} & {\bf 95.52} & {\bf 76.41} & {\bf 84.90} & {\bf 88.83} & {\bf 79.10} & {\bf 83.68} \\
    \bottomrule
    \end{tabular}%
  }
\end{table*}%

%% file: Tables/zs_blip.tex
\begin{table}[ht]
    \centering
    \caption{Zero-shot evaluation of Zero-Shot Ensemble (ZS$_{En}$) with BLIP, BLIP2 and CLIP-ViT-B/32 on ImageNet.}
    \label{tab:zs_blip}
    \resizebox{0.5\textwidth}{!}{
    \begin{tabular}{l|c}
    \toprule
        Method & Acc (\%) \\
        \midrule
        Zero-shot BLIP \cite{li2022blip} & 49.14 \\
        Zero-shot BLIP2 \cite{li2023blip} & 36.41 \\
        Zero-shot CLIP-ViT-B/32 \cite{radford2021learning} & 62.04 \\
        \midrule
        (BLIP, BLIP2) with Mean & 51.39 \\
        (BLIP, BLIP2) with ZS$_{En}$ & \textbf{56.38} \\
        \midrule
        (BLIP, BLIP2, CLIP-ViT-B/32) with Mean & 65.44 \\
        (BLIP, BLIP2, CLIP-ViT-B/32) with ZS$_{En}$ & \textbf{67.88} \\
    \bottomrule
    \end{tabular}}
\end{table}

%% file: Tables/t_blip.tex
\begin{table}[ht]
    \centering
    \caption{Base-to-new generalization of Training-free Ensemble (TF$_{En}$) and Tuning Ensemble (T$_{En}$) with BLIP, BLIP2 and CLIP-ViT-B/32 on ImageNet.}
    \label{tab:t_blip}
    \resizebox{0.8\textwidth}{!}{
    \begin{tabular}{l|ccc}
    \toprule
        Method & Base Acc (\%) & New Acc (\%) & HM Acc (\%) \\
        \midrule
        Zero-shot BLIP \cite{li2022blip} & 50.54 & 55.75 & 53.02 \\
        Zero-shot BLIP2 \cite{li2023blip} & 29.41 & 51.98 & 37.57 \\
        Zero-shot CLIP-ViT-B/32 \cite{radford2021learning} & 67.45 & 64.02 & 65.69 \\
        \midrule
        (BLIP, BLIP2) with Mean & 50.47 & 61.36 & 55.38 \\
        (BLIP, BLIP2) with TF$_{En}$ & 52.65 & 63.23 & 57.46 \\
        (BLIP, BLIP2) with T$_{En}$ & \textbf{54.74} & \textbf{66.83} & \textbf{60.18} \\
        \midrule
        (BLIP, BLIP2, CLIP-ViT-B/32) with Mean & 68.86 & 70.00 & 69.43 \\
        (BLIP, BLIP2, CLIP-ViT-B/32) with TF$_{En}$ & 70.10 & 71.50 & 70.79 \\
        (BLIP, BLIP2, CLIP-ViT-B/32) with T$_{En}$ & \textbf{71.04} & \textbf{73.92} & \textbf{72.45} \\
    \bottomrule
    \end{tabular}}
\end{table}

%% file: Tables/para.tex
\begin{table}[ht]
    \centering
    \caption{Parameter comparison among various methods.}
    \label{tab:para}
    \resizebox{0.8\textwidth}{!}{
    \begin{tabular}{c|cccccc}
    \toprule
         Method & CoOp & CoCoOp & MaPLe & PromptSRC & CoCoOp + T$_{En}$ & PromptSRC + T$_{En}$ \\
         \midrule
         \# Param & 2048 & 34K & 3.55M & 46K & 336K & 292K  \\
         Acc (\%) & 71.66 & 75.60 & 78.55 & 79.54 & \textbf{79.20} & \textbf{81.11} \\
    \bottomrule
    \end{tabular}}
\end{table}

%% file: Tables/weight_tf.tex
\begin{table}[ht]
    \centering
    \caption{The weights searched by Training-Free Ensemble (TF$_{En}$) on diverse datasets.}
    \label{tab:weight_tf}
    \resizebox{0.8\textwidth}{!}{
    \begin{tabular}{l|ccc}
    \toprule
         Dataset & W$_{RN50}$ (ZS Acc \%) & W$_{RN101}$ (ZS Acc \%) & W$_{ViT-B/32}$ (ZS Acc \%)  \\
         \midrule
         ImageNet & 0.1 (58.23) & 0.5 ( 61.26) & 0.4 (62.04) \\
         Caltech101 & 0.1 (85.96) & 0.3 (89.66) & \textbf{0.4} (91.12) \\
         OxfordPets & 0.4 (85.80) & 0.4 (86.89) & \textbf{0.6} (87.49) \\
         StanfordCars & 0.3 (55.57) & \textbf{0.8} (63.16) & 0.4 (60.37) \\
         Flowers102 & 0.3 (65.98) & 0.8 (63.95) & \textbf{0.9} (66.95) \\
         Food101 & 0.2 (77.32) & \textbf{0.3} (80.54) & 0.2 (80.47) \\
         FGVCAircraft & 0.1(17.16) & 0.7 (18.18) & 0.2 (19.20) \\
         SUN397 & 0.4 (58.53) & 0.1 (58.96) & \textbf{0.7} (62.00) \\
         DTD & 0.6 (42.38) & 0.6 (38.48) & \textbf{0.5} (43.79) \\
         EuroSAT & 0.8 (37.42) & 0.5 (32.62) & 0.1 (45.12) \\
         UCF101 & 0.1 (61.43) & 0.5 (61.04) & \textbf{0.8} (62.07) \\
    \bottomrule
    \end{tabular}}
\end{table}

%% file: main.bbl
\begin{thebibliography}{64}
\providecommand{\natexlab}[1]{#1}
\providecommand{\url}[1]{\texttt{#1}}
\expandafter\ifx\csname urlstyle\endcsname\relax
  \providecommand{\doi}[1]{doi: #1}\else
  \providecommand{\doi}{doi: \begingroup \urlstyle{rm}\Url}\fi

\bibitem[Anderson et~al.(2018)Anderson, He, Buehler, Teney, Johnson, Gould, and Zhang]{anderson2018bottom}
Anderson, P., He, X., Buehler, C., Teney, D., Johnson, M., Gould, S., and Zhang, L.
\newblock Bottom-up and top-down attention for image captioning and visual question answering.
\newblock In \emph{CVPR}, pp.\  6077--6086, 2018.

\bibitem[Bossard et~al.(2014)Bossard, Guillaumin, and Gool]{bossard2014food}
Bossard, L., Guillaumin, M., and Gool, L.~V.
\newblock Food-101--mining discriminative components with random forests.
\newblock In \emph{ECCV}, pp.\  446--461. Springer, 2014.

\bibitem[Breiman(1996)]{breiman1996bagging}
Breiman, L.
\newblock Bagging predictors.
\newblock \emph{Machine learning}, 24:\penalty0 123--140, 1996.

\bibitem[Breiman(2001)]{breiman2001random}
Breiman, L.
\newblock Random forests.
\newblock \emph{Machine learning}, 45:\penalty0 5--32, 2001.

\bibitem[Buja \& Stuetzle(2000)Buja and Stuetzle]{buja2000smoothing}
Buja, A. and Stuetzle, W.
\newblock Smoothing effects of bagging.
\newblock \emph{Preprint. AT\&T Labs-Research}, 2000.

\bibitem[Chao et~al.(2016)Chao, Changpinyo, Gong, and Sha]{chao2016empirical}
Chao, W.-L., Changpinyo, S., Gong, B., and Sha, F.
\newblock An empirical study and analysis of generalized zero-shot learning for object recognition in the wild.
\newblock In \emph{ECCV}, pp.\  52--68. Springer, 2016.

\bibitem[Chen et~al.(2024)Chen, Li, Shen, Yang, Li, Keutzer, Darrell, and Liu]{chen2024large}
Chen, L., Li, B., Shen, S., Yang, J., Li, C., Keutzer, K., Darrell, T., and Liu, Z.
\newblock Large language models are visual reasoning coordinators.
\newblock \emph{NeurIPS}, 36, 2024.

\bibitem[Cimpoi et~al.(2014)Cimpoi, Maji, Kokkinos, Mohamed, and Vedaldi]{cimpoi2014describing}
Cimpoi, M., Maji, S., Kokkinos, I., Mohamed, S., and Vedaldi, A.
\newblock Describing textures in the wild.
\newblock In \emph{CVPR}, pp.\  3606--3613, 2014.

\bibitem[Coates \& Ng(2011)Coates and Ng]{coates2011importance}
Coates, A. and Ng, A.~Y.
\newblock The importance of encoding versus training with sparse coding and vector quantization.
\newblock In \emph{ICML}, 2011.

\bibitem[Cortes et~al.(2014)Cortes, Mohri, and Syed]{cortes2014deep}
Cortes, C., Mohri, M., and Syed, U.
\newblock Deep boosting.
\newblock In \emph{ICML}, pp.\  1179--1187. PMLR, 2014.

\bibitem[Deng et~al.(2009)Deng, Dong, Socher, Li, Li, and Fei-Fei]{deng2009imagenet}
Deng, J., Dong, W., Socher, R., Li, L.-J., Li, K., and Fei-Fei, L.
\newblock Imagenet: A large-scale hierarchical image database.
\newblock In \emph{CVPR}, pp.\  248--255. Ieee, 2009.

\bibitem[Deng et~al.(2012)Deng, Yu, and Platt]{deng2012scalable}
Deng, L., Yu, D., and Platt, J.
\newblock Scalable stacking and learning for building deep architectures.
\newblock In \emph{ICASSP}, pp.\  2133--2136. IEEE, 2012.

\bibitem[Desai \& Johnson(2021)Desai and Johnson]{desai2021virtex}
Desai, K. and Johnson, J.
\newblock Virtex: Learning visual representations from textual annotations.
\newblock In \emph{CVPR}, pp.\  11162--11173, 2021.

\bibitem[Dosovitskiy et~al.(2020)Dosovitskiy, Beyer, Kolesnikov, Weissenborn, Zhai, Unterthiner, Dehghani, Minderer, Heigold, Gelly, et~al.]{dosovitskiy2020image}
Dosovitskiy, A., Beyer, L., Kolesnikov, A., Weissenborn, D., Zhai, X., Unterthiner, T., Dehghani, M., Minderer, M., Heigold, G., Gelly, S., et~al.
\newblock An image is worth 16x16 words: Transformers for image recognition at scale.
\newblock In \emph{ICLR}, 2020.

\bibitem[Dosovitskiy et~al.(2021)Dosovitskiy, Beyer, Kolesnikov, Weissenborn, Zhai, Unterthiner, Dehghani, Minderer, Heigold, Gelly, Uszkoreit, and Houlsby]{dosovitskiy2020vit}
Dosovitskiy, A., Beyer, L., Kolesnikov, A., Weissenborn, D., Zhai, X., Unterthiner, T., Dehghani, M., Minderer, M., Heigold, G., Gelly, S., Uszkoreit, J., and Houlsby, N.
\newblock An image is worth 16x16 words: Transformers for image recognition at scale.
\newblock \emph{ICLR}, 2021.

\bibitem[Elhoseiny et~al.(2013)Elhoseiny, Saleh, and Elgammal]{elhoseiny2013write}
Elhoseiny, M., Saleh, B., and Elgammal, A.
\newblock Write a classifier: Zero-shot learning using purely textual descriptions.
\newblock In \emph{ICCV}, pp.\  2584--2591, 2013.

\bibitem[Fei-Fei et~al.(2004)Fei-Fei, Fergus, and Perona]{fei2004learning}
Fei-Fei, L., Fergus, R., and Perona, P.
\newblock Learning generative visual models from few training examples: An incremental bayesian approach tested on 101 object categories.
\newblock In \emph{CVPRW}, pp.\  178--178. IEEE, 2004.

\bibitem[Freund et~al.(1996)Freund, Schapire, et~al.]{freund1996experiments}
Freund, Y., Schapire, R.~E., et~al.
\newblock Experiments with a new boosting algorithm.
\newblock In \emph{ICML}, volume~96, pp.\  148--156. Citeseer, 1996.

\bibitem[Friedman(2001)]{friedman2001greedy}
Friedman, J.~H.
\newblock Greedy function approximation: a gradient boosting machine.
\newblock \emph{Annals of statistics}, pp.\  1189--1232, 2001.

\bibitem[Frome et~al.(2013)Frome, Corrado, Shlens, Bengio, Dean, Ranzato, and Mikolov]{frome2013devise}
Frome, A., Corrado, G.~S., Shlens, J., Bengio, S., Dean, J., Ranzato, M., and Mikolov, T.
\newblock Devise: A deep visual-semantic embedding model.
\newblock \emph{NeurIPS}, 26, 2013.

\bibitem[Gomez et~al.(2017)Gomez, Patel, Rusinol, Karatzas, and Jawahar]{gomez2017self}
Gomez, L., Patel, Y., Rusinol, M., Karatzas, D., and Jawahar, C.
\newblock Self-supervised learning of visual features through embedding images into text topic spaces.
\newblock In \emph{CVPR}, pp.\  4230--4239, 2017.

\bibitem[Ha et~al.(2005)Ha, Cho, and MacLachlan]{ha2005response}
Ha, K., Cho, S., and MacLachlan, D.
\newblock Response models based on bagging neural networks.
\newblock \emph{Journal of Interactive Marketing}, 19\penalty0 (1):\penalty0 17--30, 2005.

\bibitem[He et~al.(2016)He, Zhang, Ren, and Sun]{he2016deep}
He, K., Zhang, X., Ren, S., and Sun, J.
\newblock Deep residual learning for image recognition.
\newblock In \emph{CVPR}, pp.\  770--778, 2016.

\bibitem[Helber et~al.(2019)Helber, Bischke, Dengel, and Borth]{helber2019eurosat}
Helber, P., Bischke, B., Dengel, A., and Borth, D.
\newblock Eurosat: A novel dataset and deep learning benchmark for land use and land cover classification.
\newblock \emph{IEEE Journal of Selected Topics in Applied Earth Observations and Remote Sensing}, 12\penalty0 (7):\penalty0 2217--2226, 2019.

\bibitem[Huynh \& Elhamifar(2020)Huynh and Elhamifar]{huynh2020fine}
Huynh, D. and Elhamifar, E.
\newblock Fine-grained generalized zero-shot learning via dense attribute-based attention.
\newblock In \emph{CVPR}, pp.\  4483--4493, 2020.

\bibitem[Jia et~al.(2021)Jia, Yang, Xia, Chen, Parekh, Pham, Le, Sung, Li, and Duerig]{jia2021scaling}
Jia, C., Yang, Y., Xia, Y., Chen, Y.-T., Parekh, Z., Pham, H., Le, Q., Sung, Y.-H., Li, Z., and Duerig, T.
\newblock Scaling up visual and vision-language representation learning with noisy text supervision.
\newblock In \emph{ICML}, pp.\  4904--4916. PMLR, 2021.

\bibitem[Jiang et~al.(2020)Jiang, Xu, Araki, and Neubig]{jiang2020can}
Jiang, Z., Xu, F.~F., Araki, J., and Neubig, G.
\newblock How can we know what language models know?
\newblock \emph{ACL}, 8:\penalty0 423--438, 2020.

\bibitem[Joulin et~al.(2016)Joulin, Maaten, Jabri, and Vasilache]{joulin2016learning}
Joulin, A., Maaten, L. v.~d., Jabri, A., and Vasilache, N.
\newblock Learning visual features from large weakly supervised data.
\newblock In \emph{ECCV}, pp.\  67--84. Springer, 2016.

\bibitem[Kang et~al.(2020)Kang, Chen, Quackenbush, and Ding]{kang2020novel}
Kang, T., Chen, P., Quackenbush, J., and Ding, W.
\newblock A novel deep learning model by stacking conditional restricted boltzmann machine and deep neural network.
\newblock In \emph{SIGKDD}, pp.\  1316--1324, 2020.

\bibitem[Khattak et~al.(2023{\natexlab{a}})Khattak, Rasheed, Maaz, Khan, and Khan]{khattak2023maple}
Khattak, M.~U., Rasheed, H., Maaz, M., Khan, S., and Khan, F.~S.
\newblock Maple: Multi-modal prompt learning.
\newblock In \emph{CVPR}, pp.\  19113--19122, 2023{\natexlab{a}}.

\bibitem[Khattak et~al.(2023{\natexlab{b}})Khattak, Wasim, Naseer, Khan, Yang, and Khan]{khattak2023self}
Khattak, M.~U., Wasim, S.~T., Naseer, M., Khan, S., Yang, M.-H., and Khan, F.~S.
\newblock Self-regulating prompts: Foundational model adaptation without forgetting.
\newblock In \emph{ICCV}, pp.\  15190--15200, 2023{\natexlab{b}}.

\bibitem[Krause et~al.(2013)Krause, Stark, Deng, and Fei-Fei]{krause20133d}
Krause, J., Stark, M., Deng, J., and Fei-Fei, L.
\newblock 3d object representations for fine-grained categorization.
\newblock In \emph{ICCVW}, pp.\  554--561, 2013.

\bibitem[LeBlanc \& Tibshirani(1996)LeBlanc and Tibshirani]{leblanc1996combining}
LeBlanc, M. and Tibshirani, R.
\newblock Combining estimates in regression and classification.
\newblock \emph{Journal of the American Statistical Association}, 91\penalty0 (436):\penalty0 1641--1650, 1996.

\bibitem[Lei~Ba et~al.(2015)Lei~Ba, Swersky, Fidler, et~al.]{lei2015predicting}
Lei~Ba, J., Swersky, K., Fidler, S., et~al.
\newblock Predicting deep zero-shot convolutional neural networks using textual descriptions.
\newblock In \emph{ICCV}, pp.\  4247--4255, 2015.

\bibitem[Li et~al.(2017)Li, Jabri, Joulin, and Van Der~Maaten]{li2017learning}
Li, A., Jabri, A., Joulin, A., and Van Der~Maaten, L.
\newblock Learning visual n-grams from web data.
\newblock In \emph{ICCV}, pp.\  4183--4192, 2017.

\bibitem[Li et~al.(2022)Li, Li, Xiong, and Hoi]{li2022blip}
Li, J., Li, D., Xiong, C., and Hoi, S.
\newblock Blip: Bootstrapping language-image pre-training for unified vision-language understanding and generation.
\newblock In \emph{ICML}, pp.\  12888--12900. PMLR, 2022.

\bibitem[Li et~al.(2023{\natexlab{a}})Li, Li, Savarese, and Hoi]{li2023blip}
Li, J., Li, D., Savarese, S., and Hoi, S.
\newblock Blip-2: Bootstrapping language-image pre-training with frozen image encoders and large language models.
\newblock In \emph{ICML}, pp.\  19730--19742. PMLR, 2023{\natexlab{a}}.

\bibitem[Li et~al.(2023{\natexlab{b}})Li, Lian, Lu, Bai, Chen, and Wang]{li2023graphadapter}
Li, X., Lian, D., Lu, Z., Bai, J., Chen, Z., and Wang, X.
\newblock Graphadapter: Tuning vision-language models with dual knowledge graph.
\newblock \emph{NeurIPS}, 36, 2023{\natexlab{b}}.

\bibitem[Li et~al.(2021)Li, Liang, Zhao, Cui, Ouyang, Shao, Yu, and Yan]{li2021supervision}
Li, Y., Liang, F., Zhao, L., Cui, Y., Ouyang, W., Shao, J., Yu, F., and Yan, J.
\newblock Supervision exists everywhere: A data efficient contrastive language-image pre-training paradigm.
\newblock \emph{arXiv preprint arXiv:2110.05208}, 2021.

\bibitem[Lu et~al.(2022)Lu, Liu, Zhang, Liu, and Tian]{lu2022prompt}
Lu, Y., Liu, J., Zhang, Y., Liu, Y., and Tian, X.
\newblock Prompt distribution learning.
\newblock In \emph{CVPR}, pp.\  5206--5215, 2022.

\bibitem[Lu et~al.(2023)Lu, He, Li, Song, and Xiang]{lu2023prediction}
Lu, Z., He, S., Li, D., Song, Y.-Z., and Xiang, T.
\newblock Prediction calibration for generalized few-shot semantic segmentation.
\newblock \emph{TIP}, 2023.

\bibitem[Maji et~al.(2013)Maji, Rahtu, Kannala, Blaschko, and Vedaldi]{maji2013fine}
Maji, S., Rahtu, E., Kannala, J., Blaschko, M., and Vedaldi, A.
\newblock Fine-grained visual classification of aircraft.
\newblock \emph{arXiv preprint arXiv:1306.5151}, 2013.

\bibitem[Mikolov et~al.(2013{\natexlab{a}})Mikolov, Chen, Corrado, and Dean]{mikolov2013efficient}
Mikolov, T., Chen, K., Corrado, G., and Dean, J.
\newblock Efficient estimation of word representations in vector space.
\newblock In \emph{ICLR}, 2013{\natexlab{a}}.

\bibitem[Mikolov et~al.(2013{\natexlab{b}})Mikolov, Sutskever, Chen, Corrado, and Dean]{mikolov2013distributed}
Mikolov, T., Sutskever, I., Chen, K., Corrado, G.~S., and Dean, J.
\newblock Distributed representations of words and phrases and their compositionality.
\newblock \emph{NeurIPS}, 26, 2013{\natexlab{b}}.

\bibitem[Nilsback \& Zisserman(2008)Nilsback and Zisserman]{nilsback2008automated}
Nilsback, M.-E. and Zisserman, A.
\newblock Automated flower classification over a large number of classes.
\newblock In \emph{ICCVGIP}, pp.\  722--729. IEEE, 2008.

\bibitem[Parkhi et~al.(2012)Parkhi, Vedaldi, Zisserman, and Jawahar]{parkhi2012cats}
Parkhi, O.~M., Vedaldi, A., Zisserman, A., and Jawahar, C.
\newblock Cats and dogs.
\newblock In \emph{CVPR}, pp.\  3498--3505. IEEE, 2012.

\bibitem[Radford et~al.(2021)Radford, Kim, Hallacy, Ramesh, Goh, Agarwal, Sastry, Askell, Mishkin, Clark, et~al.]{radford2021learning}
Radford, A., Kim, J.~W., Hallacy, C., Ramesh, A., Goh, G., Agarwal, S., Sastry, G., Askell, A., Mishkin, P., Clark, J., et~al.
\newblock Learning transferable visual models from natural language supervision.
\newblock In \emph{ICML}, pp.\  8748--8763. PMLR, 2021.

\bibitem[Sariyildiz et~al.(2020)Sariyildiz, Perez, and Larlus]{sariyildiz2020learning}
Sariyildiz, M.~B., Perez, J., and Larlus, D.
\newblock Learning visual representations with caption annotations.
\newblock In \emph{ECCV}, pp.\  153--170. Springer, 2020.

\bibitem[Shin et~al.(2020)Shin, Razeghi, Logan~IV, Wallace, and Singh]{shin2020autoprompt}
Shin, T., Razeghi, Y., Logan~IV, R.~L., Wallace, E., and Singh, S.
\newblock Autoprompt: Eliciting knowledge from language models with automatically generated prompts.
\newblock In \emph{EMNLP}, 2020.

\bibitem[Socher et~al.(2013)Socher, Ganjoo, Manning, and Ng]{socher2013zero}
Socher, R., Ganjoo, M., Manning, C.~D., and Ng, A.
\newblock Zero-shot learning through cross-modal transfer.
\newblock \emph{NeurIPS}, 26, 2013.

\bibitem[Soomro et~al.(2012)Soomro, Zamir, and Shah]{soomro2012ucf101}
Soomro, K., Zamir, A.~R., and Shah, M.
\newblock Ucf101: A dataset of 101 human actions classes from videos in the wild.
\newblock \emph{arXiv preprint arXiv:1212.0402}, 2012.

\bibitem[Torresani et~al.(2010)Torresani, Szummer, and Fitzgibbon]{torresani2010efficient}
Torresani, L., Szummer, M., and Fitzgibbon, A.
\newblock Efficient object category recognition using classemes.
\newblock In \emph{ECCV}, pp.\  776--789. Springer, 2010.

\bibitem[Touvron et~al.(2021)Touvron, Cord, Douze, Massa, Sablayrolles, and J{\'e}gou]{touvron2021training}
Touvron, H., Cord, M., Douze, M., Massa, F., Sablayrolles, A., and J{\'e}gou, H.
\newblock Training data-efficient image transformers \& distillation through attention.
\newblock In \emph{ICML}, pp.\  10347--10357. PMLR, 2021.

\bibitem[Vaswani et~al.(2017)Vaswani, Shazeer, Parmar, Uszkoreit, Jones, Gomez, Kaiser, and Polosukhin]{vaswani2017attention}
Vaswani, A., Shazeer, N., Parmar, N., Uszkoreit, J., Jones, L., Gomez, A.~N., Kaiser, {\L}., and Polosukhin, I.
\newblock Attention is all you need.
\newblock \emph{NeurIPS}, 30, 2017.

\bibitem[Wang et~al.(2019)Wang, Zheng, Yu, and Miao]{wang2019survey}
Wang, W., Zheng, V.~W., Yu, H., and Miao, C.
\newblock A survey of zero-shot learning: Settings, methods, and applications.
\newblock \emph{TIST}, 10\penalty0 (2):\penalty0 1--37, 2019.

\bibitem[Wang et~al.(2023)Wang, Liang, He, Xu, Wang, and Tan]{wang2023improving}
Wang, Z., Liang, J., He, R., Xu, N., Wang, Z., and Tan, T.
\newblock Improving zero-shot generalization for clip with synthesized prompts.
\newblock In \emph{CVPR}, pp.\  3032--3042, 2023.

\bibitem[Wolpert(1992)]{wolpert1992stacked}
Wolpert, D.~H.
\newblock Stacked generalization.
\newblock \emph{Neural networks}, 5\penalty0 (2):\penalty0 241--259, 1992.

\bibitem[Xian et~al.(2017)Xian, Schiele, and Akata]{xian2017zero}
Xian, Y., Schiele, B., and Akata, Z.
\newblock Zero-shot learning-the good, the bad and the ugly.
\newblock In \emph{CVPR}, pp.\  4582--4591, 2017.

\bibitem[Xiao et~al.(2010)Xiao, Hays, Ehinger, Oliva, and Torralba]{xiao2010sun}
Xiao, J., Hays, J., Ehinger, K.~A., Oliva, A., and Torralba, A.
\newblock Sun database: Large-scale scene recognition from abbey to zoo.
\newblock In \emph{CVPR}, pp.\  3485--3492. IEEE, 2010.

\bibitem[Yi et~al.(2022)Yi, Shen, Gou, and Elhoseiny]{yi2022exploring}
Yi, K., Shen, X., Gou, Y., and Elhoseiny, M.
\newblock Exploring hierarchical graph representation for large-scale zero-shot image classification.
\newblock In \emph{ECCV}, pp.\  116--132. Springer, 2022.

\bibitem[Yu et~al.(2023)Yu, Lu, Jin, Chen, and Wang]{yu2023task}
Yu, T., Lu, Z., Jin, X., Chen, Z., and Wang, X.
\newblock Task residual for tuning vision-language models.
\newblock In \emph{CVPR}, pp.\  10899--10909, 2023.

\bibitem[Zhong et~al.(2021)Zhong, Friedman, and Chen]{zhong2021factual}
Zhong, Z., Friedman, D., and Chen, D.
\newblock Factual probing is [mask]: Learning vs. learning to recall.
\newblock In \emph{NAACL}, 2021.

\bibitem[Zhou et~al.(2022{\natexlab{a}})Zhou, Yang, Loy, and Liu]{zhou2022conditional}
Zhou, K., Yang, J., Loy, C.~C., and Liu, Z.
\newblock Conditional prompt learning for vision-language models.
\newblock In \emph{CVPR}, pp.\  16816--16825, 2022{\natexlab{a}}.

\bibitem[Zhou et~al.(2022{\natexlab{b}})Zhou, Yang, Loy, and Liu]{zhou2022learning}
Zhou, K., Yang, J., Loy, C.~C., and Liu, Z.
\newblock Learning to prompt for vision-language models.
\newblock \emph{IJCV}, 130\penalty0 (9):\penalty0 2337--2348, 2022{\natexlab{b}}.

\end{thebibliography}
